\documentclass[twoside]{article}
\UseRawInputEncoding

\usepackage{enumerate}
\usepackage{amsmath}
\usepackage{amsfonts}
\usepackage[pdftex]{graphicx}
\usepackage{caption}
\usepackage{subcaption}
\usepackage{multirow}
\usepackage{float}
\usepackage{placeins}
\usepackage{booktabs}
\usepackage[accepted]{aistats2025}
\usepackage{natbib}

\bibpunct{[}{]}{,}{a}{,}{,}

\usepackage{xcolor}
\definecolor{LightBlue}{rgb}{0.3, 0., 1}

\usepackage[colorlinks=true,linkcolor=LightBlue,citecolor=LightBlue,urlcolor=LightBlue]{hyperref}

\begin{document}

%

%

\twocolumn[

\aistatstitle{Stiff Transfer Learning for Physics-Informed Neural Networks}

\aistatsauthor{ Emilien Seiler* \And Wanzhou Lei \And  Pavlos Protopapas}

\aistatsaddress{EPFL \And  Harvard University \And Harvard University} ]

\begin{abstract}
Stiff differential equations are prevalent in various scientific domains, posing significant challenges due to the disparate time scales of their components. As computational power grows, physics-informed neural networks (PINNs) have led to significant improvements in modeling physical processes described by differential equations. Despite their promising outcomes, vanilla PINNs face limitations when dealing with stiff systems, known as “failure modes”. In response, we propose a novel approach, stiff transfer learning for physics-informed neural networks (STL-PINNs), to effectively tackle stiff ordinary differential equations (ODEs) and partial differential equations (PDEs). Our methodology involves training a Multi-Head-PINN in a low-stiff regime, and obtaining the final solution in a high stiff regime by transfer learning. This addresses the failure modes related to stiffness in PINNs while maintaining computational efficiency by computing “one-shot" solutions. The proposed approach demonstrates superior accuracy and speed compared to PINNs-based methods, as well as comparable computational efficiency with implicit numerical methods in solving stiff-parameterized linear and polynomial nonlinear ODEs and PDEs under stiff conditions. Furthermore, we demonstrate the scalability of such an approach and the superior speed it offers for simulations involving initial conditions and forcing function reparametrization.
\end{abstract}

\section{Introduction}

\label{sec:intro}
Stiff differential equations present a unique challenge in numerical analysis due to their disparate time scales, where some of their components evolve rapidly while others change slowly. 
This phenomenon can produce rapid transient phases or highly oscillatory solutions when solving stiff ordinary differential equations (ODEs) and stiff partial differential equations (PDEs). These equations are prevalent in several scientific and engineering domains such as chemistry \citep{SANTILLANA2016372}, circuit simulation \citep{8316975} or astrophysics \citep{Saijo_2018}. Explicit numerical schemes face difficulties with stiffness due to the necessity of employing exceedingly small step sizes for stability \citep{book}. In contrast, implicit methods such as Radau \citep{HAIRER199993} or Runge Kutta fourth-order EDT \citep{stiffpde} are stiffness-accurate methods and provide robust solutions by effectively capturing the stiff dynamics. Although these methods have proven to be successful, they have limitations. For example, they can be quite expensive in terms of computational time, being affected by the so-called "curse of dimensionality" in high-dimensional PDEs \citep{cursePDE}.

As computational power grows and machine learning techniques evolve, researchers are turning to data-driven and learning-based approaches to solve PDEs and ODEs. One notable architecture that has recently gained attention is the so-called physics-informed neural networks (PINNs) \citep{pinns}. These models use neural networks and gradient-based optimization algorithms, leveraging automatic differentiation to enforce physical constraints. PINNs have been successfully applied to numerous important applications in the field of computational science, including computational fluid dynamics \citep{JIN2021109951}, quantum mechanics \citep{shah2022physicsinformed}, cardiac
electrophysiology simulation \citep{cardiacpinns}, and material science \citep{materialPINNs}, to name a few. Compared to numerical methods that need to interpolate and store dense meshes, PINNs offer the advantages of rapid evaluations through forward passes and only need to store the model's weights. Moreover, PINNs are an alternative when dealing with the "curse of dimensionality" \citep{cho2023separable}.

Despite the advantages and promising outcomes of PINNs, they face a fundamental limitation regarding stiff systems, known as a "failure mode" \citep{krishnapriyan2021characterizing}. \cite{wang2020understanding} identified a first failure mode linked to numerical stiffness, resulting in imbalanced back-propagated gradients during training. Then, \cite{Ji_2021} demonstrated that stiffness primarily causes PINNs to fail in simulating stiff kinetic systems. To address this issue, \cite{baty2023solving} proposed straightforward methods to enhance the training process, while \cite{pinnsformer} introduced PINNsFormer, a state-of-the-art Transformer-based framework with multi-head attention and a Wavelet activation function.

However, the previous solutions share a fundamental limitation with PINNs which hinders their applicability to a broader range of applications: it is necessary to train a new neural network from scratch to accommodate a novel parameter set of the equation (e.g., novel initial conditions, forcing function, etc.). Hence, the use of PINNs to solve scenarios that involve numerous queries is often computationally demanding. For example, in design optimization and uncertainty propagation, the parameterized ODEs and PDEs must be simulated thousands of times. As a solution, \cite{cho2023hypernetworkbased} proposed lightweight low-rank PINNs. On the other hand, our contribution is based on transfer learning.

Transfer learning (TL) represents a potential solution to the problem of computing parametrized solutions without the need for retraining. The architecture of multi-head physics-informed neural networks (MH-PINN) enables to learn the "latent space" of a class of differential equations \citep{MHPINNs}. From this, \cite{desai2022oneshot} proposed a transfer learning framework for PINNs, which led to "one-shot" inference for linear systems of PDEs and ODEs. This implies that highly accurate solutions for numerous unknown differential equations can be obtained instantaneously without retraining the entire network. This transfer learning approach has already been applied to the simulation of branched flows \citep{pellegrin2022transfer}. \cite{lei2023oneshot} extended this approach to nonlinear ODEs and PDEs with polynomial terms using the perturbation method.

We propose stiff transfer learning for physics-informed neural
networks (STL-PINNs), a method tailored to tackle stiffness in ODEs and PDEs. This method unfolds i) training a MH-PINN in a low-stiff regime of the equation, and ii) transfer learning to compute the solution in a high-stiff regime.\\ In essence, this method addresses the two critical challenges mentioned before i) PINNs "failure mode" from stiffness during training, since the base model is trained in a low-stiff regime, and ii) the computational burden of retraining PINNs when changing parameters, as final solutions are computed in "one-shot” transfer learning.\\
We show the efficacy and efficiency of our proposed method by solving stiff-parametrized linear and nonlinear ODEs and PDEs, such as \emph{duffing equation} or \emph{advection-reaction equation}. Our contributions include:
\begin{enumerate}
\item 
We develop a method based on transfer learning to solve stiff equations by training the model in a least-stiff regime.
\item We demonstrate that the proposed method outperforms state-of-the-art PINNs-based methods in both accuracy and speed, in "failure modes" associated with stiffness.
\item We show that our method's computational time is independent of stiffness and faster than the implicit methods for the given examples.
\item We establish that our method is scalable regarding training regimes.
\item We show that the proposed method offers significantly superior speed compared to numerical methods for simulations involving initial conditions or forcing function reparametrization.
\end{enumerate}

\section{Preliminary Components}

We are interested in linear systems of ODEs and PDEs of the following form:
\begin{equation} \label{eq:1}
\sum_{j=1}^{d} C_j \frac{\partial y}{\partial x_j} + B \frac{\partial y}{\partial t} + A y = f(\boldsymbol{x}) 
\end{equation}

The PDE in Eq.\ref{eq:1} is characterized by the solution \( y \in \mathbb{R}^n \). The vector \( \boldsymbol{x} = [t, x_1, \ldots, x_d] \in \mathbb{R}^{d+1} \) contains \(t \in \mathbb{R}\) the independent variable of time and \( d \) the independent variables of spatiality. \( A, B, C_k \in \mathbb{R}^{n \times n} \) are coefficient matrices and \( f: \mathbb{R}^{d+1} \xrightarrow{} \mathbb{R}^{n} \) denotes the forcing function. In the subsequent section, the general PDE notation is employed and can be applied to ODEs by setting \(d=0\), \(\boldsymbol{x}=t\) and \(\frac{\partial}{\partial t} = \frac{d}{dt}\).

\subsection{Multi-Head PINNs}
\label{subsec:MH-PINNs}

\begin{figure*}[!h]
  \centering
  \includegraphics[width=0.8\linewidth]{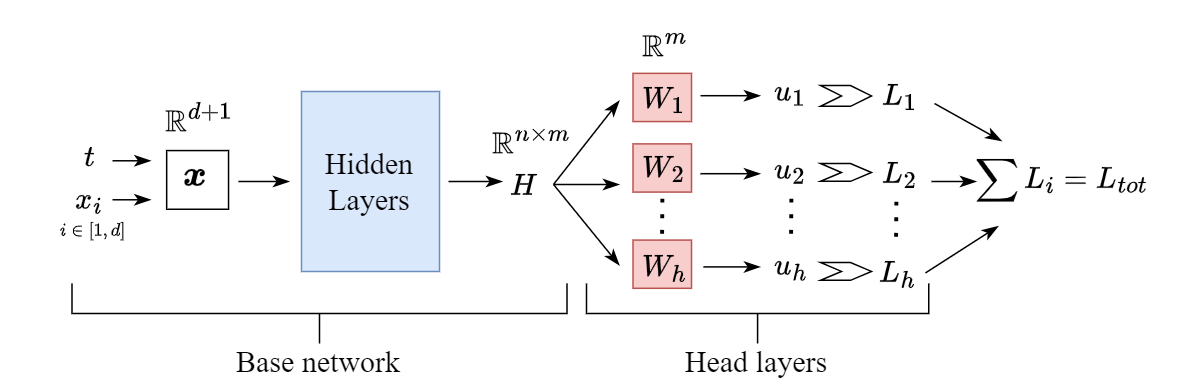}
  \caption{MH-PINNs architecture with \(h\) heads. The base network constructs \(H \in \mathbb{R}^{n \times m}\) from input \(\boldsymbol{x}\) that passes through multiple hidden layers. \(m\) is the dimension of the last hidden layer and \(n\) is the dimension of the system of equations. The \(h\) head layers consist of separate linear layers \(W_i\) without any activation function. Each of them produces individual outputs \(u_i=HW_i\) and their loss. These losses are then summed up to yield to the final loss \(L_{tot}\).}
  \label{fig:1}
\end{figure*}

The idea of the MH-PINNs architecture (Fig.\ref{fig:1}) is to approximate equations of the same form over different parameters. It comprises multiple heads with distinct sets of parameters (e.g., equation coefficients, forcing functions, or initial conditions). Each head is associated with a unique set of weights, \(W_i\), which are used to generate outputs \(u_i= HW_i\), using the same \(H\). Hence, it encodes the generalized form of the equation in the latent representation \(H\). The following loss function is applied to each head output:\begin{equation*} \label{eq:2}
\begin{split}
L = & \frac{\omega_{1}}{\!N_1}\sum^{N_1}_{i=1}\!\left(\sum_{j=1}^{d}C_j\!\left(\!\frac{\partial u}{\partial x_j}\!\right)_{\!\!\!\boldsymbol{x_i}}\!\!\!\!\! + \!B\!\left(\!\frac{\partial u}{\partial t}\!\right)_{\!\!\!\boldsymbol{x_i}}\!\!\!\!\! + \!Au_{\boldsymbol{x_i}} \!\!- \!f_{\boldsymbol{x_i}}\!\right)^{\!\!\!2}\\
& + \frac{\omega_{2}}{\!N_2}\sum^{N_2}_{i=1}\left(u_{\boldsymbol{x_i^{b}}} \!- y_{\boldsymbol{x_i^{b}}}\right)^{\!2}
\end{split}
\tag{2}
\end{equation*}
\setcounter{equation}{2}where \(u\) is the neural network approximation of the true solution \(y\). \(N_1\) is the number of collocation points and \(\boldsymbol{x_i}\) the \(i^{th}\) one. \(N_2\) is the number of boundary or initial conditions collocation points and \(\boldsymbol{x^{b}_i}\) the \(i^{th}\) one. The notation \(\left(\cdot\right)_{\boldsymbol{x}}\) indicates the evaluation of the function at the point \(\boldsymbol{x}\). The residual and boundary conditions are leveraged by weights \(\omega_1\) and \(\omega_2\). While training, the network adjusts \(W_i\) to suit different parameter sets and converges to a fix \(H\).

\subsection{One-Shot Transfer Learning}
\label{subsec:OSTL}
After training with MH-PINNs, the final value \(H\) is saved as \(H_{fix}\). A head layer can be considered as a matrix multiplication operation \({u}=H_{fix}W\). Then, for untrained heads with different parameters, namely, \(\hat{C_j}\), \(\hat{B}\), \(\hat{A}\), \(\hat{f}\), \(\hat{y}_{\boldsymbol{x_i^b}}\) only \(W\) computation is needed.
The optimal weight, denoted as \(\hat{W}\), is derived by minimizing the loss function of Eq.\ref{eq:2} after substitution of \(u=H_{fix}W\). The following Eq.\ref{eq:3} is the solution of \(\nabla_W L=0\). Here, \(H_{fix}\) is denoted as \(H\) for clarity (see App.\ref{app:A} for derivation details):
\begin{equation}\label{eq:3} \begin{split}
&\hat{W} \!= \!M^{-1}\!\left( \frac{\omega_1}{\!N_1}\sum_{i=1}^{N_1}{H_{\boldsymbol{x_i}}^{*}}^{\!\!T} \hat{f}_{\boldsymbol{x_i}} + \frac{\omega_2}{\!N_2}\sum_{i=1}^{N_2}H_{\boldsymbol{\!x_i^{b}}}^{T} \hat{y}_{\boldsymbol{x_i^{b}}}\right)\\
&M \!= \!\frac{\omega_1}{\!N_1}\sum_{i=1}^{N_1}{H_{\boldsymbol{x_i}}^{*}}^{\!\!T} H_{\boldsymbol{x_i}}^{*} + \frac{\omega_2}{\!N_2}\sum_{i=1}^{N_2}H_{\boldsymbol{\!x_i^{b}}}^{T} H_{\boldsymbol{\!x_i^{b}}}\\
&H^{*}_{\boldsymbol{x_i}} \!= \!\sum_{j=1}^{d}\hat{C_j} H_{\boldsymbol{x_i}}^{x_j} + \hat{B} H_{\boldsymbol{x_i}}^{t} + \hat{A}H_{\boldsymbol{x_i}}\\
&\text{with } H^{\left(\cdot\right)} 
 \text{ is the derivative}
 \text{ with respect to } t, x_j
\end{split}
\end{equation}
Hence, we can compute the solution of a new parameter set \(u = H_{fix}\hat{W}\). In practice, transfer learning to new coefficients \(\hat{C_j}\), \(\hat{B}\) and \(\hat{A}\) requires the inversion of the \(M\) matrix. However, this is not necessary for \(\hat{f}\) and \(\hat{y}_{\boldsymbol{x_i^b}}\) because they are independent of \(M\). Consequently, transfer learning to new forcing functions, initial conditions, or boundary conditions requires less computational time.

\begin{figure*}[ht]
  \centering
  \includegraphics[width=0.9\linewidth]{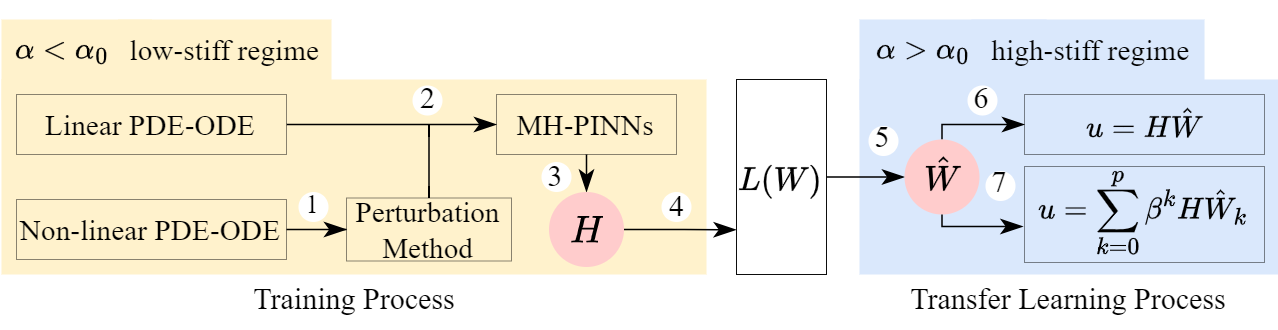}
  \caption{
   STL-PINN diagram process:
   \\1. Linearize non-linear PDE-ODE. \hfill 4. Substitute \(u=HW\) in the loss function. \hspace*{0.52cm}
   \\2. Train MH-PINN in a low-stiff regime. \hfill 5. Compute \(\hat{W}\) in a high-stiff regime with Eq.\ref{eq:3}. 
   \\ 3. Extract and fix \(H\) after training.    \hfill6\&7. Compute final solution \(u\). \hspace*{2.4cm}
   }
  \label{fig:2}
\end{figure*}

\subsection{Perturbation Method}
\label{subsec:PM}
Nonlinear differential equations do not possess an analytical solution for \(W\). For nonlinearity with polynomial terms, we consider ODEs and PDEs of the following form:
\begin{equation} \label{eq:4}
\sum_{j=1}^{d} C_j \frac{\partial y}{\partial x_j} + B \frac{\partial y}{\partial t} + A y + \beta g(y)= f(\boldsymbol{x}) 
\end{equation}
with \(\beta \in [0, 1]\) and \(g:\mathbb{R}^n \xrightarrow{} \mathbb{R}^n\) a nonlinear polynomial function.
To remove the nonlinearity in the equation, we
approximate the nonlinear term \(\beta{g(y)}\)
as a composition of functions using perturbation expansion \citep{kevorkian1981perturbation}. Assuming \(y=\sum_{k=0}^{\infty}\beta^{k}Y_k\),
where \(Y_{k}\) are unknown functions of \(\boldsymbol{x}\), we approximate \(y\) with only \(p\) terms as \(y \approx \sum_{k=0}^{p} \beta^{k}Y_k\). After substitution in Eq.\ref{eq:4} and by collecting the power of \(\beta\), the following formula provides an approximation of the original nonlinear equation (see App.\ref{app:B} for derivation details):
\begin{equation} \label{eq:5}
\sum_{j=1}^{d} C_j \frac{\partial Y_k}{\partial x_j} + B \frac{\partial Y_k}{\partial t} + A Y_k = F_k\left(\boldsymbol{x}, Y_0, ..., Y_{k-1}\right) 
\end{equation}
\(\forall k \in \mathbb{N}, ~0 \leq k \leq p\), the forcing function \(F_k\) depends only on previously solved \(Y_0, Y_1, Y_{k-1}\). Therefore, Eq.\ref{eq:4} is reduced to a series of \(p \!+\! 1\) linear differential equations of the same form.\\
To perform transfer learning for nonlinear ODEs and PDEs such as Eq.\ref{eq:4}, we first trained a MH-PINNs on its linear counterpart Eq.\ref{eq:5} with \(k=0\). Subsequently, we can use Eq.\ref{eq:3} to calculate \(\hat{W}_k\), solving iteratively the series of \(p\!+\!1\) linear differential equations, and reconstruct the nonlinear solution \(u = \sum_{k=0}^{p}\beta^kH_{\text{fix}}\hat{W}_k\).

\subsection{Stiffness Characterization}
The term "phenomenon" seems more appropriate than "property," as the latter suggests that stiffness can be precisely defined mathematically, which, unfortunately, is proving difficult. The following statements are used to characterize stiffness:
\paragraph{Stiffness Ratio (SR):}
Eigenvalues of the Jacobian matrix associated with the system of ODEs that cover a wide range of magnitudes are indicative of stiffness. The following stiffness ratio \(SR = max|\lambda|/min|\lambda|\) is introduced. If \(SR < 20\) the problem is not stiff, from \(SR \approx 1000\) the problem is classified as stiff \citep{inbook}.

\paragraph{Transient Phase:} Stiffness creates rapid transient phases where the rate of change of the solution is significantly higher compared to other time scales \citep{book}.

\paragraph{Restricted Stability Regions for Explicit Methods:} Stiffness requires very small step sizes for explicit numerical methods to remain stable, making them computationally inefficient \citep{book}.

Stiffness in a differential equation can arise based on specific initial conditions, parameters, or regions within the problem space. Although Jacobian eigenvalues
certainly contribute to the stiffness of the equation, they often cannot be the only criterion for
characterizing stiffness \citep{Higham1993}. While SR serves as an indicator of stiffness
within an equation, it cannot be extrapolated for comparison between equations because it only captures the effect of the eigenvalues. Quantities such as the dimension of the system, the initial conditions, the forcing function, the smoothness of the solution, or the integration interval must also be considered. 

In this work, these three statements are governed by a parameter \(\alpha\). As \(\alpha\) increases, the equation becomes stiffer, the SR value increases, the transient phase becomes more rapid, and explicit methods require a longer time to obtain a solution.

\section{STL-PINN: Stiff Transfer Learning for Physics-Informed Neural Network}

The general idea of the STL-PINN method is to solve stiff equations by transfer learning from a low-stiff regime to a high-stiff regime. Fig.\ref{fig:2} explains the overall process of solving stiff-parameterized linear and nonlinear PDEs-ODEs. The key feature of our approach is that the training process is conducted in a low-stiff regime where \(\alpha < \alpha_0\), and the final solution is then computed in a high-stiff regime where \(\alpha > \alpha_0\). The training heads operate in low-stiff regimes governed by distinct \(\alpha\) values. Hence, MH-PINN's latent representation \(H\) captures variations resulting from changes in the \(\alpha\) value, enabling transfer to high-stiff regimes. Moreover, distinct initial conditions or forcing functions can be used on the training heads. This allows the latent representation \(H\) to capture variations arising from these changes. After the training process, transfer learning can be applied multiple times across different stiffness regimes and various equation parameter sets (e.g., initial conditions, forcing functions), without the need for retraining. Finally, STL-PINN overcomes the "failure mode" of PINNs associated with training under stiff conditions, and the final solution is computed in "one-shot", making the method computationally efficient.

\section{Experiments}
We demonstrate that our method significantly outperforms baselines on stiff-parameterized 2-dimensional linear, nonlinear ODEs, and 2-dimensional linear PDE.

\paragraph{Baselines for comparison}The objective of this study is to compare the STL-PINN method to vanilla PINN \citep{pinns} and the state-of-the-art PINNsFormer \citep{pinnsformer} models in terms of accuracy and computational time. Additionally, the study aims to compare STL-PINNs with numerical methods, specifically explicit RK45 \citep{BUTCHER1996247} and implicit Radau method \citep{HAIRER199993}, in terms of computational time.

\paragraph{Accuracy}  For accuracy metrics, we first employ the \(L_2\) relative error, defined as \(L_2^{rel}= \sum_{i=1}^{N}\sqrt{(u_{\boldsymbol{x_i}} - y_{\boldsymbol{x_i}})^2}/\sum_{i=1}^{N}\sqrt{y_{\boldsymbol{x_i}}^2}\) being the PINN solution and \(y_{\boldsymbol{x_i}}\) the Radau solution evaluated at \(\boldsymbol{x_i}\). It offers a measure of the overall performance of the method across all time scales. We also use the \(L_1\) relative error define as \(L_1^{rel}=\sum_{i=1}^{N}|u_{\boldsymbol{x_i}} - y_{\boldsymbol{x_i}}|/\sum_{i=1}^{N}|y_{\boldsymbol{x_i}}|\). Finally, we use the \(L_{\infty}\) maximum relative error defines as \( \L_{\infty}^{rel} = \text{max}_{i \in [1, N]}
|u_{\boldsymbol{x_i}} - y_{\boldsymbol{x_i}}|/\text{max}_{i \in [1, N]}
|y_{\boldsymbol{x_i}}|\), which serves as an important metric, enhancing the error magnitude caused by rapid transient phases that can occur on a very short time scale.

\paragraph{Computational time} STL-PINN objective is to compute multiple solutions of an equation through post-training transfer learning. When discussing computational time, we consider the wall-clock time during the transfer learning phase, after the latent representation \(H\) has been exported from training.

\subsection{Stiff-parameterized ODEs-PDEs}
The following equations are derived from their general representation available in App.\ref{app:E}.
\paragraph{2D linear ODEs:} For our first stiff-parameterized linear 2-dim ODE, we consider the following overdamped-harmonic-oscillator (OHO) equation:
\begin{equation}\label{eq:6}
    \begin{cases}
        \frac{dy_1}{dt} - y_2 = 0 \!\!\!\!\quad \quad \quad \quad \, y_1\!\left(t\!=\!0\right) \!=\! y_{1_0}\\
        \frac{dy_2}{dt} + y_1 + \alpha y_2 = 0\!\!\!\!\quad \, \, y_2\!\left(t\!=\!0\right) \!=\!  y_{2_0}
    \end{cases}\!\!\!\!\!,\!\!\!\!\quad  t\in[0, T],
\end{equation}
where \(y_1\) describes the motion of a harmonic oscillator slowed by a frictional force proportional to \(y_2\), the velocity of the motion. The stiffness parameter \(\alpha\) governs the stiffness of the equation and SR is proportional to \(\alpha^2\). The range of \(\alpha>30\) poses a significant challenge in training vanilla PINNs due to the rapid transient phase in the vicinity of \(y_2\left(t=0\right)\) caused by the stiffness.

For our second stiff-parameterized linear 2-dim ODE, we consider the following linear ODEs with non-constant forcing function (NCFF) \citep{NCFF}:
\begin{equation*}\label{eq:7}
    \begin{split}
        \begin{cases}
            \frac{dy_1}{dt} + 2y_1 - y_2 = 2\sin\!\left(\omega t\right)\\
            \frac{dy_2}{dt} + \left(1-\alpha\right) y_1 + \alpha y_2 = \alpha\!\left(\cos\!\left(\omega t\right)\!-\!\sin\!\left(\omega t\right)\right)\\
            \text{with } y_1\!\left(t\!=\!0\right) \!=\!  y_{1_0}\text{, } y_2\!\left(t\!=\!0\right) \!=\!  y_{2_0}
        \end{cases}
    \end{split}
    \tag{7}
\end{equation*}\setcounter{equation}{7}where \(y_1\), \(y_2\) describe an oscillation motion governed by the angular frequency \(\omega\) with a rapid transient phase in the vicinity of \(t=0\). SR is proportional to $\alpha$ and values of \(\alpha>30\) pose a significant challenge in training vanilla PINNs.

\paragraph{2D nonlinear ODE:} For a stiff-parameterized nonlinear polynomial 2-dim ODE, we consider the following Duffing equation:
\begin{equation}\label{eq:8}
    \begin{cases}
        \frac{dy_1}{dt} - y_2 = 0\\
        \frac{dy_2}{dt} + y_1 + \alpha y_2 + \beta y_1^3 = cos\!\left(t\right)\\
        \text{with } y_1\!\left(t\!=\!0\right) \!=\!  y_{1_0}\text{, } y_2\!\left(t\!=\!0\right) \!=\!  y_{2_0}
    \end{cases}
\end{equation}
with \(\beta \!\in ]0, 1[\) and \(y_1\) describes the motion of a damped oscillator with a more complex potential due to a periodic driving force and the \(\beta\) term that controls the amount of nonlinearity in the restoring force. SR is proportional to \(\alpha^2\). For \(\alpha>40\), vanilla PINNs training faces challenges due to rapid transient phases near \(y_2\!\left(t\!=\!0\right)\).

\paragraph{2D linear PDE:} For a stiff-parameterized PDE we consider the 2-dim linear advection-reaction equation (AR):
\begin{equation}\label{eq:9}
    \begin{cases}
        \frac{dy_1}{dt} + \mu\frac{dy_1}{dx} + \alpha\left(k_1 y_1 - k_2 y_2\right) = 0\\
        \frac{dy_2}{dt} + \mu\frac{dy_2}{dx} - \alpha\left(k_1 y_1 - k_2 y_2\right) = 0\\
        t\in[0, T], x\in[0, L],
    \end{cases} 
\end{equation}
with \(y_1\!\left(t, x\!=\!0\right)\!=\!y_1\!\left(t, x\!=\!L\right) \!=\! 0\),\\ 
\hspace*{0.75cm}\(y_2\!\left(t, x\!=\!0\right)\!=\!y_2\!\left(t, x\!=\!L\right) \!=\!0\),\\\hspace*{0.8cm}\!\(y_1\!\left(t\!=\!0, x\right)\!=\!y_{1_0}\!\!\left(x\right)\), \!\(y_2\!\left(t\!=\!0, x\right) =y_{2_0}\!\!\left(x\right)\), clamped \hspace*{0.6cm} cubic spline with maximum value \(y_0^{max}\).\\
This system comes from the advection-reaction atmospheric equation. It is characterized by the evolution of chemical species in a constant wind field with stiff chemistry mechanisms. As investigated by \cite{SANTILLANA2016372}, the stiffness comes from the reaction dynamics being much faster than the advection one. The parameters are set to \(\mu\!=\!1.8\!\times\! 10^{-4}\), \(k_1\!=\!1\!\times\!10^{-3}\), \(k_2\!=\!2\!\times\!10^{-3}\) with effective stiffness of the order \(\mathcal{O}(10^{-1})\). The stiffness is governed by \(\alpha\) that changes the dynamics of the chemical reaction. For \(\alpha>30\) vanilla PINNs struggle because of a rapid transient phase when the chemical components reach equilibrium. For numerical simulations, we used Godunov splitting and implemented an explicit second-order accurate (in space) method based on the LaxWendroff (LW) scheme with superbee slope limiters \citep{LeVeque_2002}. For the chemistry, we used the implicit Radau method.

\subsection{Experimental setup}

\paragraph{Performance analysis} We aim to investigate the performance of STL-PINNs in terms of accuracy and computational time.  To achieve this, Table \ref{tab:1} presents the training head configurations of MH-PINN in a low-stiff regime of the ODEs and PDEs. In the case of the Duffing equation, the choice of the perturbation parameter \(p\) is optimized within the range \(p \in [0, 20]\), as detailed in App.\ref{app:F}. Training setup including optimization, loss and model parameters are available in App.\ref{app:G}.

\begin{table}[h]
  \caption{MH-PINN training heads}\label{tab:1}
  \centering
  \renewcommand{\arraystretch}{0.5}
  \resizebox{0.9\linewidth}{!}{%

  \begin{tabular}{@{}l@{}ccccccccccc}
  \toprule
    & \textbf{Head} & 1 & 2& 3& 4& 5& 6& 7& 8& 9& 10\\
    \midrule
    \multirow{2}{*}{\textbf{OHO}} &     \(\alpha\) & 2 & 4 & 6 & 8 & 10 & 12 & 14 & 16 & 18 & 20 \\
    & \(y_0\) & \multicolumn{10}{c}{[1, 0.5] for all heads} \\
    \midrule
    \multirow{3}{*}{\textbf{NCFF}} &     \(\alpha\) & 2 & 4 & 6 & 8 & 10 & 12 & 14 & 16 & 18 & 20 \\
    & \(y_0\) & \multicolumn{10}{c}{[2, 4] for all heads} \\
    &     \(\omega\) & \multicolumn{10}{c}{1 for all heads} \\
    \midrule
    \multirow{3}{*}{\textbf{Duffing}} &     \(\alpha\) & 13 & 16 & 19 & 22 & 25 & 28 & 31 & 34 & 37 & 40 \\
    & \(y_0\) & \multicolumn{10}{c}{[1, 0.5] for all heads}\\
    &     \(\beta\) & \multicolumn{10}{c}{0.5 for all heads} \\
    \midrule
    \multirow{2}{*}{\textbf{AR}} &     \(\alpha\) & 3 & 6 & 8 & 12 & - & - & - & - & - & - \\
    & \(y_0^{max}\!\!\!\!\) & 1 & 1 & 1 & 1 & - & - & - & - & - & -\\
    \bottomrule 
  \end{tabular}}
\end{table}

\paragraph{Scalability Analysis}
We aim to investigate the scalability of STL-PINN in the stiff transfer regime relative to the training one. To achieve this, we trained multiple MH-PINNs with increasing ranges of \(\alpha\) values up to \(\alpha_{max}\). Then, we analyze the resulting variations in transfer learning accuracy. Training head configurations are available in App.\ref{app:H}.

\paragraph{Reparametrization Analysis} We aim to evaluate the performance of STL-PINN in the reparametrization of initial conditions or forcing functions within a stiff regime. These components do not necessitate the inversion of the \(M\) matrix (see Eq.\ref{eq:3}) during the transfer learning phase, thereby offering efficient computational times. To conduct this analysis, we trained a MH-PINN in a low-stiff regime with various initial conditions or forcing functions to capture their behaviors. Then, we conducted 1000 simulations of transfer learning in a high-stiff regime by varying these parameters and reported the average time and accuracy metrics. For OHO, Duffing, and AR equations, the initial conditions vary within specific ranges: \(y_{1_0} \!\! \in \! [0, 5]\) and \(y_{2_0} \!\! \in \! [0, 5]\) for OHO, \(y_{1_0} \!\! \in \! [0.5, 2]\) and \(y_{2_0} \!\! \in \! [0, 1]\) for Duffing, \(y_0^{max} \!\! \in \! [0.5, 2]\) for AR. On the other hand, the forcing function varies within the interval  \(\omega \! \in \! [0, \pi]\) for the NCFF. Training head configurations are available in App.\ref{app:I}.

\subsection{Experimental results}

\begin{figure*}[t]
\centering
\begin{subfigure}{0.6\textwidth}
  \centering
  \includegraphics[width=0.9\linewidth]{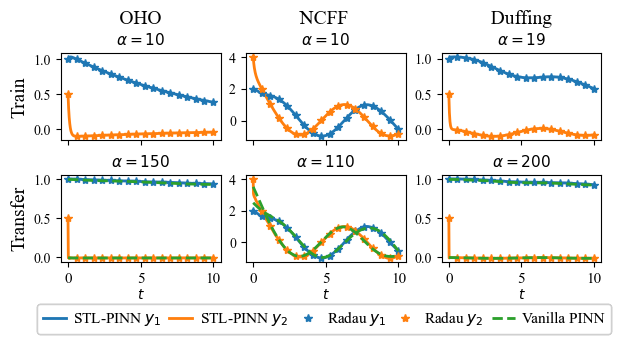}
  \caption{ODEs: STL-PINN and Radau}
  \label{fig:3sub1}
\end{subfigure}%
\begin{subfigure}{0.4\textwidth}
  \centering
  \includegraphics[width=1\linewidth]{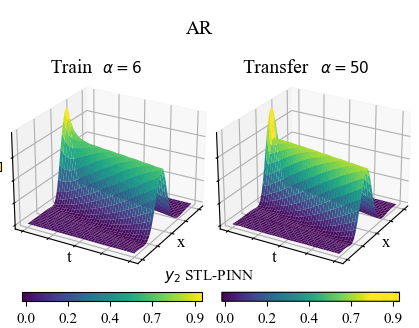} 
  \caption{PDE: STL-PINN}
  \label{fig:3sub2}
\end{subfigure}
\caption{Results for (a) OHO, NCFF, Duffing ODEs (b) AR PDE. For each equation, an example of both training low-stiff and transfer high-stiff STL-PINN solutions is shown. Low-stiff regimes take \(\alpha = \{10, 10, 19, 6\}\) while high-stiff regimes take \(\alpha = \{150, 110, 200, 50\}\) for OHO, NCFF, Duffing, AR respectively. Additionally for ODEs, the Radau solution is plotted  as an accurate approximation and the vanilla PINN solution is provided for the high-stiff regimes.}
\label{fig:3}
\end{figure*}
 
\begin{figure}[h]
  \centering
\includegraphics[width=1\linewidth]{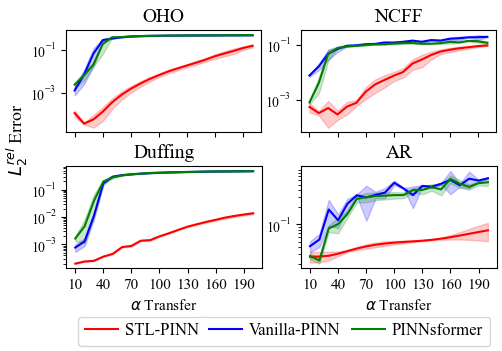}
  \caption{\(L_2\) relative error of vanilla PINN, PINNsFormer and STL-PINN solving stiff-parameterized ODEs-PDEs as stiffness regime increases, with \(\alpha \in [10, 200]\). The reported errors are the average of three separate runs, along with 90\% confidence intervals.} 
  \label{fig:new}
\end{figure}

\paragraph{Performance analysis: Accuracy}  Fig.\ref{fig:3} shows examples of training and transfer solutions. As expected, training solutions are well aligned with Radau solutions, as they are computed in a low-stiff regime. Moreover, STL-PINNs solutions computed by transfer learning in a high-stiff regime are also accurate. In contrast to vanilla PINNs, our method effectively captures stiff behavior, as evidenced by its precise encoding of sharp transient phases in the vicinity of initial conditions. Vanilla PINNs do not adequately capture these rapid transient phases because of the "failure mode" caused by stiffness. More figures about training and transfer learning solutions are provided in App.\ref{app:J}.

Fig.\ref{fig:new} compares STL-PINN, vanilla PINN, and PINNsFormer. Vanilla PINN and PINNsFormer fail to encode stiffness, with significant errors appearing at $\alpha = 30$ for OHO, $\alpha = 20$ for NCFF, $\alpha = 40$ for Duffing, and $\alpha = 30$ for AR. STL-PINN consistently outperforms these methods across all $\alpha$ values, with performance improvement of at least an order of magnitude, even in high-stiffness regimes. While STL-PINN’s error increases with stiffness, it remains more accurate than other PINNs. Thus, we demonstrate that the STL-PINN method can capture stiffness and obtain accurate solutions over "failure modes" of PINN. The scalability analysis further discusses the limits of transfer learning, and additional metrics are in App.\ref{app:more_mertics}.

\paragraph{Performance analysis: Computational time} Table \ref{tab:3} compares the wall-clock time of STL-PINN with numerical methods. As expected, explicit RK45's computational time significantly increases from \(10^{-2}\) to \(10^{-1}\) seconds as \(\alpha\) rises from 20 to 200. In contrast, STL-PINN and Radau maintain consistent computational times regardless of the equation's stiffness. For ODEs, STL-PINN is faster than Radau, with times in the range of \(10^{-3}\) against \(10^{-2}\) seconds. For the AR equation, our method is significantly faster than LW-Radau, with times in the range of \(10^{-2}\) against \(10^{1}\) seconds. 

\paragraph{Scalability Analysis} Fig.\ref{fig:5} illustrates that with a larger training range for 
, STL-PINN can transfer to higher stiff regimes with reduced error. Indeed, across all equations, larger \(\alpha_{max}\) results in a lower error curve and enables transfer learning to stiffer regimes while maintaining the same error magnitude. For example, with the Duffing equation, we reach \(10^{-4}\) error at \(\alpha\!=\!60\) with \(\alpha_{max}\!=\!20\). Then, doubling \(\alpha_{max}\) to \(40\), we can now transfer to \(\alpha\!=\!140\) with the same \(10^{-4}\) error. More metrics are available in App.\ref{app:M}. This demonstrates the scalability of our method: the more we train in stiff regimes, the more we extend the range of transfer learning to even stiffer regimes.
\begin{figure}[h]
  \centering  \includegraphics[width=0.95\linewidth]{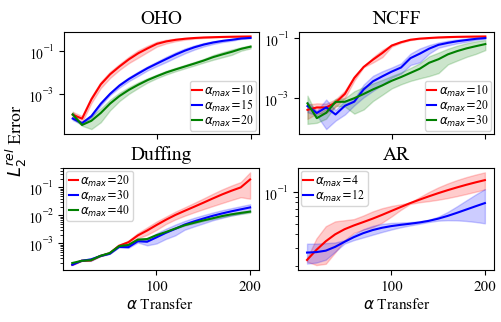}
  \caption{\(L_2\) relative error of STL-PINN over increasing ranges of \(\alpha\) during training, where \(\alpha_{max}\) denotes the maximum value. The x-axis is the \(\alpha \) transfer range. The reported errors are the average of three separate runs, along with 90\% confidence intervals.}
  \label{fig:4}
\end{figure}

\begin{table*}[t]
  \caption{Average wall-clock time over 100 runs of STL-PINN, RK45, Radau and LW-Radau methods solving stiff-parametrized ODEs-PDEs as stiffness regime increases, \(\alpha \in [10, 200]\).}
  \label{tab:3}
  \centering
  \resizebox{0.95\linewidth}{!}{%
  \begin{tabular}{lccccccccccc}
  \toprule
    \multicolumn{1}{c}{} & \multicolumn{3}{c}{\textbf{OHO}} & \multicolumn{3}{c}{\textbf{NCFF}}& \multicolumn{3}{c}{\textbf{Duffing}} & \multicolumn{2}{c}{\textbf{AR}}
    \\
    \cmidrule(lr){2-4}\cmidrule(lr){5-7}\cmidrule(lr){8-10}\cmidrule(lr){11-12}
    {\(\boldsymbol{\alpha}\)} & \textbf{STL-PINN} & \textbf{Radau}  & \textbf{RK45}  & \textbf{STL-PINN} & \textbf{Radau}  & \textbf{RK45} & \textbf{STL-PINN} & \textbf{Radau}  & \textbf{RK45}  & \textbf{STL-PINN} & \textbf{LW-Radau}\\
    \cmidrule(lr){2-12}
     & \multicolumn{3}{c}{\textit{Time sec.}} &\multicolumn{3}{c}{\textit{Time sec.}} &\multicolumn{3}{c}{\textit{Time sec.}} &\multicolumn{2}{c}{\textit{Time sec.}}\\
     \midrule
    20 & \(0.0033\) & \(0.014\) & \(0.026\) & \(0.0026\) & \(0.011\) & \(0.017\) & \(0.0041\) & \(0.012\) & \(0.016\) & \(0.046\) & \(57\)\\
    50 & \(0.0030\) & \(0.016\) & \(0.028\) & \(0.0026\) & \(0.013\) & \(0.034\) & \(0.0038\) & \(0.010\) & \(0.031\) & \(0.046\) & \(57\)\\
    100 & \(0.0032\) & \(0.020\) & \(0.051\) & \(0.0029\) & \(0.011\) & \(0.063\) & \(0.0042\) & \(0.010\) & \(0.057\) & \(0.046\) & \(57\)\\
    150 & \(0.0035\) & \(0.017\) & \(0.073\) & \(0.0029\) & \(0.012\) & \(0.092\) & \(0.0033\) & \(0.010\) & \(0.089\) & \(0.046\) & \(58\)\\
    200 & \(0.0031\) & \(0.018\) & \(0.092\) & \(0.0025\) & \(0.013\) & \(0.11\) & \(0.0029\) & \(0.010\) & \(0.11\) & \(0.046\) & \(58\)\\
\bottomrule 
      \end{tabular}}
\end{table*}

\newpage
\paragraph{Reparametrization Analysis} Training outcomes on the reparametrization analysis are available in App.\ref{app:K}. Table \ref{tab:4} displays the time and accuracy of the reparametrization analysis. Here, STL-PINN performed reparametrization accurately, with mean absolute error comparable to the one involving only a change in stiffness regime. Additionally, we observe that for ODEs, the wall-clock time order of STL-PINN is \(10^{-4}\) seconds. Then, our method is faster than both RK45 and Radau by three and two magnitude orders respectively. Regarding the AR equation, STL-PINN is notably faster than the LW-Radau method,  with wall-clock time smaller than four magnitude orders.

\begin{table}[!h]
  \caption{Average wall-clock time solving 1000 simulations in a high-stiff regime, involving initial conditions or forcing function reparametrization. The mean absolute error of STL-PINN is also provided. Please note that LW-Radau is used for symbol \(^*\).}
  \label{tab:4}
  \centering
  \resizebox{0.9\linewidth}{!}{%
  \begin{tabular}{lccccc}
  \toprule
    & \textbf{OHO} & \textbf{NCFF}& \textbf{Duffing} & \textbf{AR} \\
    \midrule
    {\(\boldsymbol{\alpha}\)} & 150 & 110 & 150 & 100\\
    \midrule
    \multicolumn{1}{c}{} & 
    \multicolumn{4}{c}{\textit{Time sec.}}\\
    \cmidrule(lr){2-5}
    \textbf{STL-PINN}& 0.00015 & 0.00081 & 0.0031 & 0.0052\\
    \textbf{Radau}& 0.016 & 0.014 & 0.010 & \multirow{2}{*}{\(58^*\)}\\
    \textbf{RK45}& 0.091 & 0.076 & 0.091\\
    \cmidrule(lr){2-5}
    \textbf{MAE}& \(2.3 \times 10^{-3}\) & \(1.2 \times 10^{-2}\) & \(6.5 \times 10^{-4}\) & \(1.6 \times 10^{-2}\) \\
\bottomrule 
  \end{tabular}}
\end{table}

\section{Limitations and Future Works}
\label{sec:limitation}
We applied STL-PINN using the perturbation method to a nonlinear example, the Duffing equation. However, in some stiff problems, such as the van der Pol equation \citep{vdp}, there is no finite perturbation solution. To address this, instead of computing the final layer \(W\) in "one-shot", we could approximate it using an optimization algorithm such as LBFGS \citep{NIPS2016_8ebda540}. Since the base network \(H\) remains fixed, optimizing only the last layer would produce relatively rapid evaluations.

Although the need to linearize nonlinear PDEs introduces potential errors, complicating the assessment of the method's effectiveness.

We can compute accurate solutions in stiff regimes, challenges arise with extremely stiff regimes, such as \( \alpha\!\!=\!\!1000 \). Scalability analysis shows that the more we train in stiff regimes, the more we extend the range of transfer learning to even stiffer regimes. Therefore, for very high-stiff regimes, we need a model capable of training within an already stiff regime. One solution might involve modifying the MH-PINN architecture or training process to mitigate the stiffness "failure mode" \citep{wang2020understanding}, enabling transfer to very high-stiff regimes.

Our STL-PINN mainly concentrates on cases of stiffness involving a rapid transient phase. In future studies, our framework could be extended to highly oscillatory solutions caused by stiffness.

\section{Conclusion}
We proposed STL-PINN, a simple yet powerful method that employs transfer learning to solve stiff-parameterized ODEs and PDEs. Experimental results showed that our method can solve stiff equations with high precision, even in PINN's "failure modes". The present method outperform state-of-the-art PINNsFormer method in term of accuracy, while being faster than numerical methods. These performance results pave the way for the application of our method in experiments involving numerous simulations. We believe that this work highlights the potential of PINNs transfer learning in various scientific applications.

\newpage
\bibliography{reference}

\begin{thebibliography}{}

\bibitem[Anne~Kværnø, 2020]{NCFF}
Anne~Kværnø, M.~G. (2020).
\newblock Stiff differential equations.

\bibitem[Baty, 2023]{baty2023solving}
Baty, H. (2023).
\newblock Solving stiff ordinary differential equations using physics informed neural networks (pinns): simple recipes to improve training of vanilla-pinns.

\bibitem[Berahas et~al., 2016]{NIPS2016_8ebda540}
Berahas, A.~S., Nocedal, J., and Takac, M. (2016).
\newblock A multi-batch l-bfgs method for machine learning.
\newblock In Lee, D., Sugiyama, M., Luxburg, U., Guyon, I., and Garnett, R., editors, {\em Advances in Neural Information Processing Systems}, volume~29. Curran Associates, Inc.

\bibitem[Brugnano et~al., 2009]{vdp}
Brugnano, L., Mazzia, F., and Trigiante, D. (2009).
\newblock Fifty years of stiffness.
\newblock {\em Recent Advances in Computational and Applied Mathematics}.

\bibitem[Butcher, 1996]{BUTCHER1996247}
Butcher, J. (1996).
\newblock A history of runge-kutta methods.
\newblock {\em Applied Numerical Mathematics}, 20(3):247--260.

\bibitem[Cho et~al., 2023a]{cho2023separable}
Cho, J., Nam, S., Yang, H., Yun, S.-B., Hong, Y., and Park, E. (2023a).
\newblock Separable physics-informed neural networks.

\bibitem[Cho et~al., 2023b]{cho2023hypernetworkbased}
Cho, W., Lee, K., Rim, D., and Park, N. (2023b).
\newblock Hypernetwork-based meta-learning for low-rank physics-informed neural networks.

\bibitem[Debnath and Chinthavali, 2018]{8316975}
Debnath, S. and Chinthavali, M. (2018).
\newblock Numerical-stiffness-based simulation of mixed transmission systems.
\newblock {\em IEEE Transactions on Industrial Electronics}, 65(12):9215--9224.

\bibitem[Desai et~al., 2022]{desai2022oneshot}
Desai, S., Mattheakis, M., Joy, H., Protopapas, P., and Roberts, S. (2022).
\newblock One-shot transfer learning of physics-informed neural networks.

\bibitem[Hairer and Wanner, 1996]{book}
Hairer, E. and Wanner, G. (1996).
\newblock {\em Solving Ordinary Differential Equations II. Stiff and Differential-Algebraic Problems}, volume~14.

\bibitem[Hairer and Wanner, 1999]{HAIRER199993}
Hairer, E. and Wanner, G. (1999).
\newblock Stiff differential equations solved by radau methods.
\newblock {\em Journal of Computational and Applied Mathematics}, 111(1):93--111.

\bibitem[Hammer, 1958]{cursePDE}
Hammer, P.~C. (1958).
\newblock <i>dynamic programming</i>. richard bellman. princeton university press, princeton, n.j., 1957. xxv+ 342 pp. \$6.75.
\newblock {\em Science}, 127(3304):976--976.

\bibitem[Higham and Trefethen, 1993]{Higham1993}
Higham, D.~J. and Trefethen, L.~N. (1993).
\newblock Stiffness of odes.
\newblock {\em BIT Numerical Mathematics}, 33(2):285--303.

\bibitem[Ji et~al., 2021]{Ji_2021}
Ji, W., Qiu, W., Shi, Z., Pan, S., and Deng, S. (2021).
\newblock Stiff-pinn: Physics-informed neural network for stiff chemical kinetics.
\newblock {\em The Journal of Physical Chemistry A}, 125(36):8098–8106.

\bibitem[Jin et~al., 2021]{JIN2021109951}
Jin, X., Cai, S., Li, H., and Karniadakis, G.~E. (2021).
\newblock Nsfnets (navier-stokes flow nets): Physics-informed neural networks for the incompressible navier-stokes equations.
\newblock {\em Journal of Computational Physics}, 426:109951.

\bibitem[Kassam and Trefethen, 2005]{stiffpde}
Kassam, A.-K. and Trefethen, L.~N. (2005).
\newblock Fourth-order time-stepping for stiff pdes.
\newblock {\em SIAM Journal on Scientific Computing}, 26(4):1214--1233.

\bibitem[Kevorkian and Cole, 1981]{kevorkian1981perturbation}
Kevorkian, J. and Cole, J. (1981).
\newblock {\em Perturbation Methods in Applied Mathematics}.
\newblock Applied Mathematical Sciences. Springer.

\bibitem[Krishnapriyan et~al., 2021]{krishnapriyan2021characterizing}
Krishnapriyan, A.~S., Gholami, A., Zhe, S., Kirby, R.~M., and Mahoney, M.~W. (2021).
\newblock Characterizing possible failure modes in physics-informed neural networks.

\bibitem[Lagaris et~al., 1998]{pinns}
Lagaris, I., Likas, A., and Fotiadis, D. (1998).
\newblock Artificial neural networks for solving ordinary and partial differential equations.
\newblock {\em IEEE Transactions on Neural Networks}, 9(5):987--1000.

\bibitem[Lei et~al., 2023]{lei2023oneshot}
Lei, W., Protopapas, P., and Parikh, J. (2023).
\newblock One-shot transfer learning for nonlinear odes.

\bibitem[Lepik and Hein, 2014]{inbook}
Lepik, Ã. and Hein, H. (2014).
\newblock {\em Stiff Equations}, pages 45--57.

\bibitem[LeVeque, 2002]{LeVeque_2002}
LeVeque, R.~J. (2002).
\newblock {\em High-Resolution Methods}, page 100–128.
\newblock Cambridge Texts in Applied Mathematics. Cambridge University Press.

\bibitem[Pellegrin et~al., 2022]{pellegrin2022transfer}
Pellegrin, R., Bullwinkel, B., Mattheakis, M., and Protopapas, P. (2022).
\newblock Transfer learning with physics-informed neural networks for efficient simulation of branched flows.

\bibitem[Sahli~Costabal et~al., 2020]{cardiacpinns}
Sahli~Costabal, F., Yang, Y., Perdikaris, P., Hurtado, D.~E., and Kuhl, E. (2020).
\newblock Physics-informed neural networks for cardiac activation mapping.
\newblock {\em Frontiers in Physics}, 8.

\bibitem[Saijo, 2018]{Saijo_2018}
Saijo, M. (2018).
\newblock Determining the stiffness of the equation of state using low dynamical instabilities in differentially rotating stars.
\newblock {\em Physical Review D}, 98(2).

\bibitem[Santillana et~al., 2016]{SANTILLANA2016372}
Santillana, M., Zhang, L., and Yantosca, R. (2016).
\newblock Estimating numerical errors due to operator splitting in global atmospheric chemistry models: Transport and chemistry.
\newblock {\em Journal of Computational Physics}, 305:372--386.

\bibitem[Shah et~al., 2022]{shah2022physicsinformed}
Shah, K., Stiller, P., Hoffmann, N., and Cangi, A. (2022).
\newblock Physics-informed neural networks as solvers for the time-dependent schr\"odinger equation.

\bibitem[Wang et~al., 2020]{wang2020understanding}
Wang, S., Teng, Y., and Perdikaris, P. (2020).
\newblock Understanding and mitigating gradient pathologies in physics-informed neural networks.

\bibitem[Zhang et~al., 2022]{materialPINNs}
Zhang, E., Dao, M., Karniadakis, G.~E., and Suresh, S. (2022).
\newblock Analyses of internal structures and defects in materials using physics-informed neural networks.
\newblock {\em Science Advances}, 8(7):eabk0644.

\bibitem[Zhao et~al., 2024]{pinnsformer}
Zhao, Z., Ding, X., and Prakash, B.~A. (2024).
\newblock Pinnsformer: A transformer-based framework for physics-informed neural networks.

\bibitem[Zou and Karniadakis, 2023]{MHPINNs}
Zou, Z. and Karniadakis, G. (2023).
\newblock L-hydra: Multi-head physics-informed neural networks.

\end{thebibliography}

\newpage
\onecolumn 
\appendix
\section{Derivation of One-Shot TL Weight \(\hat{W}\)}\label{app:A}
\subsection{Linear PDE General Case}
Following Section.\ref{subsec:OSTL}, after substitution of \(u=H_{fix}W\), the loss function Eq.\ref{eq:2} for an untrained head with parameters \(\hat{C_j}\), \(\hat{B}\), \(\hat{A}\), \(\hat{f}\), \(\hat{y}_{\boldsymbol{x_i^b}}\) is a quadratic function with respect to the head layer weights \(W\) (\(H_{fix}\) denoted as \(H\) for clarity).
\begin{equation}\notag
L(W) = \frac{\omega_{1}}{\!N_1}\sum^{N_1}_{i=1}
\!\left(\sum_{j=1}^{d}
\hat{C_j} H_{\boldsymbol{x_i}}^{x_j}W + \hat{B} H_{\boldsymbol{x_i}}^{t}W + \hat{A}H_{\boldsymbol{x_i}}W- \hat{f}_{\boldsymbol{x_i}}\!\right)^{\!\!\!2}\!\!+\frac{\omega_{2}}{\!N_2}\sum^{N_2}_{i=1}\left(H_{\boldsymbol{\!x_i^{b}}}W \!- \hat{y}_{\boldsymbol{x_i^{b}}}\right)^{\!2}
\end{equation}
\begin{itemize}
    \item \(N_1\) is the number of collocation points
    \item \(\boldsymbol{x_i}\) the \(i^{th}\) collocation point
    \item \(N_2\) is the number of boundary/initial conditions collocation points
    \item \(\boldsymbol{x^{b}_i}\) the \(i^{th}\) boundary/initial conditions collocation points
    \item \(\left(\cdot\right)_{\boldsymbol{x}}\) evaluation of \(\left(\cdot\right)\) at \(\boldsymbol{x}\)
    \item \(H^{\left(\cdot\right)}\) the derivative of H with respect to \(\left(\cdot\right)\)
    \item \(\omega_1\) weight of ODE / PDE loss
    \item \(\omega_2\) weight of boundary/initial conditions loss
\end{itemize}
Thus \(u_{x_i} = H_{\boldsymbol{x_i}}W\), \(\left(\!\frac{\partial u}{\partial x_j}\!\right)_{\!\!\boldsymbol{x_i}} \!\!\! = H_{\boldsymbol{x_i}}^{x_j}W\), \(\left(\!\frac{\partial u}{\partial t}\!\right)_{\!\boldsymbol{x_i}} \!\!\! = H_{\boldsymbol{x_i}}^{t}W\). 

With dimensions:
\begin{itemize}
    \item \(\boldsymbol{x_i}, \boldsymbol{x_i^{b}} \in \mathbb{R}^{d+1}\) 
    \item \(H_{\boldsymbol{x_i}}, H_{\boldsymbol{x_i}}^{x_j}, H_{\boldsymbol{x_i}}^{t} \in \mathbb{R}^{n \times m}\)
    \item \(W \in \mathbb{R}^{m}\)
    \item \(\hat{C_j}, \hat{B},\hat{A} \in \mathbb{R}^{n \times n}\)
    \item \(\hat{f}, \hat{y}_{\boldsymbol{x_i^b}} \in \mathbb{R}^{n}\)
    \item \(m\) is the dimension of the last hidden layer of the base network
    \item \(n\) the ODE / PDE system's dimension
\end{itemize}
Let's simplify the expression by introducing the variable \(H^{*}\):
\begin{equation} \notag
H^{*}_{\boldsymbol{x_i}} \!= \!\sum_{j=1}^{d}\hat{C_j} H_{\boldsymbol{x_i}}^{x_j} + \hat{B} H_{\boldsymbol{x_i}}^{t} + \hat{A}H_{\boldsymbol{x_i}}
\end{equation}

After substitution of \(H^{*}\) into the loss:
\begin{equation}\notag
L(W) = \frac{\omega_{1}}{\!N_1}\sum^{N_1}_{i=1}
\!\left(H^{*}_{\boldsymbol{x_i}}W - \hat{f}_{\boldsymbol{x_i}}\!\right)^{\!2}\!\!+\frac{\omega_{2}}{\!N_2}\sum^{N_2}_{i=1}\left(H_{\boldsymbol{\!x_i^{b}}}W \!- \hat{y}_{\boldsymbol{x_i^{b}}}\right)^{\!2}
\end{equation}

Taking the gradient of \(L\) with respect to \(W\):
\begin{equation} \notag
    \nabla_W L(W) = \frac{2\omega_{1}}{\!N_1}\sum^{N_1}_{i=1}
\!\left({H^{*}_{\boldsymbol{x_i}}}^{\!\!T} \left( H^{*}_{\boldsymbol{x_i}} W - \hat{f}_{\boldsymbol{x_i}} \!\right)\right)\!\! + \frac{2\omega_{2}}{N_2} \sum^{N_2}_{i=1} \left({H_{\boldsymbol{x_i^{b}}}}^{T} \left( H_{\boldsymbol{x_i^{b}}} W - \hat{y}_{\boldsymbol{x_i^{b}}} \right)\right)
\end{equation}

We want \(\hat{W}\) such that \(\nabla_W L(\hat{W})=0\):
\begin{equation} \notag
    \frac{\omega_{1}}{\!N_1}\sum^{N_1}_{i=1}
\!\left({H^{*}_{\boldsymbol{x_i}}}^{\!\!T} \left( H^{*}_{\boldsymbol{x_i}} \hat{W} - \hat{f}_{\boldsymbol{x_i}} \!\right)\right)\!\! + \frac{\omega_{2}}{N_2} \sum^{N_2}_{i=1} \left({H_{\boldsymbol{x_i^{b}}}}^{T} \left( H_{\boldsymbol{x_i^{b}}} \hat{W} - \hat{y}_{\boldsymbol{x_i^{b}}} \right)\right) = 0
\end{equation}

Bringing \(\hat{W}\) terms on one side of the equation:
\begin{equation} \notag
\frac{\omega_{1}}{\!N_1}\sum^{N_1}_{i=1}
\!\left({H^{*}_{\boldsymbol{x_i}}}^{\!\!T} H^{*}_{\boldsymbol{x_i}}\right)\!\hat{W} \!\! + \frac{\omega_{2}}{N_2} \sum^{N_2}_{i=1} \left({H_{\boldsymbol{x_i^{b}}}}^{T} H_{\boldsymbol{x_i^{b}}}\right) \!\hat{W} = \frac{\omega_{1}}{N_1} \sum^{N_1}_{i=1} \!\left({H^{*}_{\boldsymbol{x_i}}}^{\!\!T} \hat{f}_{\boldsymbol{x_i}}\!\right) + \frac{\omega_{2}}{N_2} \sum^{N_2}_{i=1} \left(H_{\boldsymbol{x_i^{b}}}^{T} \hat{y}_{\boldsymbol{x_i^{b}}}\right)
\end{equation}

Regrouping \(\hat{W}\) terms:
\begin{equation} \notag
\left(\frac{\omega_{1}}{\!N_1}\sum^{N_1}_{i=1}
\!\left({H^{*}_{\boldsymbol{x_i}}}^{\!\!T} H^{*}_{\boldsymbol{x_i}}\right)\! + \frac{\omega_{2}}{N_2} \sum^{N_2}_{i=1} \left({H_{\boldsymbol{x_i^{b}}}}^{T} H_{\boldsymbol{x_i^{b}}}\right)\right)\hat{W} = \frac{\omega_{1}}{N_1} \sum^{N_1}_{i=1} \!\left({H^{*}_{\boldsymbol{x_i}}}^{\!\!T} \hat{f}_{\boldsymbol{x_i}}\!\right) + \frac{\omega_{2}}{N_2} \sum^{N_2}_{i=1} \left(H_{\boldsymbol{x_i^{b}}}^{T} \hat{y}_{\boldsymbol{x_i^{b}}}\right)
\end{equation}

Introducing the variable \(M \in \mathbb{R}^{m \times m}\) :
\begin{equation} \notag
M = \frac{\omega_{1}}{\!N_1}\sum^{N_1}_{i=1}
\!\left({H^{*}_{\boldsymbol{x_i}}}^{\!\!T} H^{*}_{\boldsymbol{x_i}}\right)\! + \frac{\omega_{2}}{N_2} \sum^{N_2}_{i=1} \left({H_{\boldsymbol{x_i^{b}}}}^{T} H_{\boldsymbol{x_i^{b}}}\right)
\end{equation}

Substituting \(M\) into the equation, we get the same weight \(\hat{W}\) as Eq.\ref{eq:3}:
\begin{equation} \notag
\hat{W} = M^{-1} \left( \frac{\omega_{1}}{N_1} \sum^{N_1}_{i=1} \!\left({H^{*}_{\boldsymbol{x_i}}}^{\!\!T} \hat{f}_{\boldsymbol{x_i}}\!\right) + \frac{\omega_{2}}{N_2} \sum^{N_2}_{i=1} \left({H_{\boldsymbol{x_i^{b}}}}^{T} \hat{y}_{\boldsymbol{x_i^{b}}}\right) \right)
\end{equation}

\subsection{Linear ODE Special Case}
In an ODE, there is only one independent variable \(t\) so the variable \(H^{*}\) is:
\begin{equation} \notag
H^{*}_{t_i} \!= \hat{B} H_{{t_i}}^{t} + \hat{A}H_{t_i}
\end{equation}

There is also only one initial condition collocation point so \(N_2=1\) and \(\boldsymbol{x_i^{b}}=t_0\). Then, \(M\) is reduced to:
\begin{equation} \notag
M_{ode} = \frac{\omega_{1}}{\!N_1}\sum^{N_1}_{i=1}
\!\left({H^{*}_{t_i}}^{\!T} H^{*}_{t_i}\right)\! + \omega_{2} {H_{t_0}}^{T} H_{t_0}
\end{equation}

Thus, \(\hat{W}\) formula becomes:
\begin{equation} \notag
\hat{W}_{ode}  = M^{-1}\left(\frac{\omega_{1}}{\!N_1}\sum^{N_1}_{i=1}
\!\left({H^{*}_{t_i}}^{\!T} \hat{f}_{t_i}\!\right) + \omega_{2} {H_{t_0}}^{T} \hat{y}_{t_0}\right)
\end{equation}

\section{Derivation of Linear Systems after Perturbation Expansion}
\label{app:B}
We start with the nonlinear equation representation:
\begin{equation} \notag
\sum_{j=1}^{d} C_j \frac{\partial y}{\partial x_j} + B \frac{\partial y}{\partial t} + A y + \beta g(y)= f(\boldsymbol{x}) 
\end{equation}
with \(\beta \in [0, 1]\) and \(g:\mathbb{R}^n \xrightarrow{} \mathbb{R}^n\) a nonlinear polynomial function.

After perturbation expansion \(y \approx \sum_{k=0}^{p} \beta^{k}Y_k\):
\begin{equation} \notag
\sum_{j=1}^{d}\left( C_j \sum_{k=0}^{p} \beta^{k} \frac{\partial Y_k}{\partial x_j}\right) + B \sum_{k=0}^{p} \beta^{k} \frac{\partial Y_k}{\partial t} + A \sum_{k=0}^{p}\beta^{k} Y_k + \beta g(\sum_{k=0}^{p} \beta^{k} Y_k)= f(\boldsymbol{x}) 
\end{equation}

After substitution of \(g\) a polynomial nonlinear equation \(g(y) = y^q\):
\begin{equation} \notag
\sum_{j=1}^{d}\left( C_j \sum_{k=0}^{p} \beta^{k} \frac{\partial Y_k}{\partial x_j}\right) + B \sum_{k=0}^{p} \beta^{k} \frac{\partial Y_k}{\partial t} + A \sum_{k=0}^{p}\beta^{k} Y_k + \beta \left(\sum_{k=0}^{p} \beta^{k} Y_k\right)^q = f(\boldsymbol{x}) 
\end{equation}

Using the multinomial theorem:
\begin{equation} \notag
\left(\sum_{k=0}^{p} \beta^{k} Y_k\right)^q=\sum_{l_0 + l_1 + ... + l_p=q} \frac{q!}{l_0!l_1!...l_p!} \beta^{\sum^{p}_{n=0}nl_n} \prod_{k=0}^{p}Y_k^{l_i}
\end{equation}

Substituting into the equation:
\begin{equation} \notag
\begin{split}
\sum_{j=1}^{d}\left( C_j \sum_{k=0}^{p} \beta^{k} \frac{\partial Y_k}{\partial x_j}\right) + B \sum_{k=0}^{p} \beta^{k} \frac{\partial Y_k}{\partial t} + A \sum_{k=0}^{p}\beta^{k} Y_k + \\
\beta \left( \sum_{l_0 + l_1 + ... + l_p=q} \frac{q!}{l_0!l_1!...l_p!} \beta^{\sum^{p}_{n=0}nl_n} \prod_{k=0}^{p}Y_k^{l_i}\right) = f(\boldsymbol{x}) 
\end{split}
\end{equation}

Collecting the first power of \(\beta\) into linear systems:
\begin{equation}\notag
  \beta^0: \sum_{j=1}^{d}\left( C_j  \frac{\partial Y_0}{\partial x_j}\right) + B  \frac{\partial Y_0}{\partial t} + A Y_0 = f(\boldsymbol{x})
\end{equation}
\begin{equation}\notag
  \beta^1: \sum_{j=1}^{d}\left( C_j  \frac{\partial Y_1}{\partial x_j}\right) + B  \frac{\partial Y_1}{\partial t} + A Y_1 = -Y_0^q
\end{equation}
\begin{equation}\notag
  \beta^2: \sum_{j=1}^{d}\left( C_j  \frac{\partial Y_2}{\partial x_j}\right) + B  \frac{\partial Y_2}{\partial t} + A Y_2 = -qY_0^{q-1}Y_1
\end{equation}

Generalizing for \(\beta^k\) with k \(\in \mathbb{N}, ~0 \leq k \leq p\) gives  \(p+1\) linear systems of the form:
\begin{equation} \notag
\sum_{j=1}^{d} C_j \frac{\partial Y_k}{\partial x_j} + B \frac{\partial Y_k}{\partial t} + A Y_k = F_k\left(\boldsymbol{x}, Y_0, ..., Y_{k-1}\right)
\end{equation}
with
\begin{equation} \notag
F_0 = f\left(\boldsymbol{x}\right),  \quad
F_k = -\sum_{\substack{l_0 + l_1 + \dots + l_p=q \\ \sum_{n=0}^{p}nl_n=k-1}} \frac{q!}{l_0!l_1!...l_p!} \prod_{k=0}^{p}Y_k^{l_i}
\end{equation}
with the forcing function \(F_k\) depending on the previous \(Y_{0}, Y_{1}, Y_{k-1}\) solutions.\\The derivation of Eq.\ref{eq:5} is then presented.

\section{General Representation of ODEs and PDEs}\label{app:E}

In this section we introduce the ODEs and PDEs examples in the general form of Eq.\ref{eq:1}. As a reminder we are interested in linear systems of ODEs and PDEs of the following form:

\begin{equation} \notag
\sum_{j=1}^{d} C_j \frac{\partial y}{\partial x_j} + B \frac{\partial y}{\partial t} + A y = f(\boldsymbol{x}) 
\end{equation}

\subsection{OHO}
\[y=[y_1, y_2]^T \in \mathbb{R}^2, \quad  d=0, \quad \boldsymbol{x}=t \in \mathbb{R}\]
\[B=\begin{bmatrix}
1 & 0\\
0 & 1
\end{bmatrix},\quad
A=\begin{bmatrix}
0 & -1\\
1 & \alpha
\end{bmatrix},\quad
f(t)=\begin{bmatrix}
0\\
0
\end{bmatrix}\]

\subsection{NCFF}
\[y=[y_1, y_2]^T \in \mathbb{R}^2, \quad  d=0, \quad \boldsymbol{x}=t \in \mathbb{R}\]
\[B=\begin{bmatrix}
1 & 0\\
0 & 1
\end{bmatrix},\quad
A=\begin{bmatrix}
2 & -1\\
1-\alpha & \alpha
\end{bmatrix},\quad
f(t)=\begin{bmatrix}
2sin\left(\omega t\right)\\
\alpha\left(cos\left(\omega t\right) - sin\left(\omega t\right)\right)
\end{bmatrix}\]

\subsection{Duffing}
Because of the nonlinearity, we take Eq.\ref{eq:4} as a reference which adds the non-linear function \(g\):
\[y=[y_1, y_2]^T \in \mathbb{R}^2, \quad  d=0, \quad \boldsymbol{x}=t \in \mathbb{R}\]
\[B=\begin{bmatrix}
1 & 0\\
0 & 1
\end{bmatrix},\quad
A=\begin{bmatrix}
0 & -1\\
0.1 & \alpha
\end{bmatrix},\quad g\left(y\right)=\begin{bmatrix}
0\\
y_1^3
\end{bmatrix}, \quad
f(t)=\begin{bmatrix}
0\\
cos(t)
\end{bmatrix}\]

\subsection{AR}
\[y=[y_1, y_2]^T \in \mathbb{R}^2, \quad  d=1, \quad \boldsymbol{x}=[t, x]^T \in \mathbb{R}^2\]
\[C_1=\begin{bmatrix}
\mu & 0\\
0 & \mu
\end{bmatrix}, \quad
B=\begin{bmatrix}
1 & 0\\
0 & 1
\end{bmatrix},\quad
A=\begin{bmatrix}
\alpha k_1 & -\alpha k_2\\
-\alpha k_1 & \alpha k_2
\end{bmatrix},\quad
f(t)=\begin{bmatrix}
0\\
0
\end{bmatrix}\]

\section{Perturbation Expansion Parameter \(p\) Optimization}\label{app:F}

\begin{figure}[h]
  \centering
\includegraphics[width=0.9\linewidth]{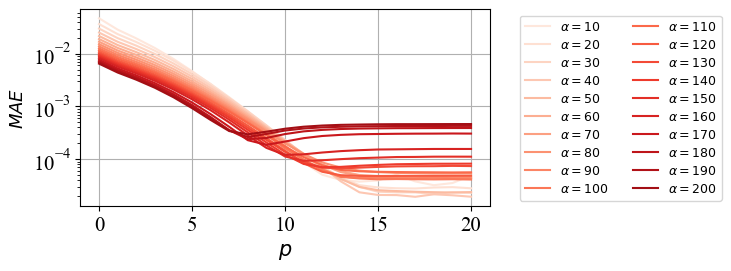}
  \caption{Mean absolute error of STL-PINN with \(p\in [0, 20]\) and \(\alpha \in [10, 200]\), solving the Duffing equation. } 
  \label{fig:5}
\end{figure}

In the case of the Duffing equation, we need to choose a perturbation parameter \(p\) since it is a nonlinear equation. For each value of \(\alpha\) in Figure \ref{fig:5}, the optimal value \(p_{opt}\) is identified as the one that minimizes the mean absolute error of \(p \in [0, 20]\).

\newpage
\section{Optimization, Loss, Model and  Compute Resources Settings}\label{app:G}

The optimization algorithm employed is Adam, with a learning rate decay of \(\gamma\) applied to each parameter group every step-size epoch. The loss function comes from Eq.\ref{eq:2} and the model is MH-PINN. The following table presents the characteristics of these components for each ODE-PDE.
\begin{table}[h]
  \caption{Training setup with optimization, loss and model settings}\notag
  \centering
  \resizebox{0.95\linewidth}{!}{%
  \begin{tabular}{lcccccccccc}
  \toprule
    & \multicolumn{4}{c}{\textbf{Optimization}} & \multicolumn{4}{c}{\textbf{Loss}} & \multicolumn{2}{c}{\textbf{MH-PINN}}\\
    \cmidrule(lr){2-5}\cmidrule(lr){6-9} \cmidrule(lr){10-11}
    & \(l_r\) & \(\gamma\) & step-size & Epoch & \(\omega_1\) & \(\omega_2\) & \(N_1\) & \(N_2\) & Layer & Activation\\
    \midrule
    \textbf{OHO} & 0.0001 & 0.98 & 100 & 20000 & 1 & 1 & 512 & 1 & [128, 128, 256] & \textit{SiLU} \\
    \textbf{NCFF} & 0.0005 & 0.98 & 100 & 40000 & 1 & 1 & 512 & 1 & [128, 128, 256] & \textit{SiLU} \\
    \textbf{Duffing} & 0.0005 & 0.98 & 100 & 40000 & 1 & 1 & 512 & 1 & [128, 128, 256] & \textit{SiLU} \\
    \textbf{AR} & 0.0001 & 0.97 & 100 & 30000 & 10000 & 1 & 9376 & 483 & [128, 128, 256, 256, 512] & \textit{SiLU} \\
    \bottomrule 
  \end{tabular}}
\end{table}
\FloatBarrier
These parameters have been chosen empirically after hyper-parameter optimization. For ODEs, we used a NVIDIA GeForce GTX 1050 with 2GB of memory, and for PDEs, we used a Tesla V100-SXM2-16GB with 16GB of memory. A bigger model is used for AR because PDE requires more complexity, and a better GPU is used to reduce training computational time. To train AR, we used a 2D grid of \(\left(75, 125\right)\) so \(N_1=75 \times 125=9376\) and used 201 collocation points for boundary conditions, 81 for initial conditions, so \(N_2 = 2*201+81 = 483\).

\section{Scalability Analysis: Training Heads}\label{app:H}
In the following section, we provide the training head configurations of the STL-PINN scalability analysis. We trained multiple MH-PINNs with increasing ranges of \(\alpha\). The training head characteristics are reported in the tables below.

\subsection{OHO}
For every model and across all heads, the initial value is set as \(y_0=[1, 0.5]\).
\begin{table}[h]
  \caption{Scalability analysis for OHO: Stiff regimes of the training heads.}\notag
  \centering
  \resizebox{0.7\linewidth}{!}{%
  \begin{tabular}{cccccccccccc}
  \toprule
    \textbf{Model} & \textbf{Head} & 1 & 2& 3& 4& 5& 6& 7& 8& 9& 10\\
    \midrule
    \(\textbf{1}\) &  \(\alpha\) & 1 & 2 & 3 & 4 & 5 & 6 & 7 & 8 & 9 & 10 \\
    \(\textbf{2}\) &     \(\alpha\) & 1 & 3 & 5 & 7 & 9 & 11 & 12 & 13 & 14 & 15 \\
    \(\textbf{3}\) &     \(\alpha\) & 2 & 4 & 6 & 8 & 10 & 12 & 14 & 16 & 18 & 20 \\
    \(\textbf{4}\) &     \(\alpha\) & 1 & 5 & 9 & 13 & 15 & 17 & 19 & 21 & 23 & 25 \\
    \bottomrule 
  \end{tabular}}
\end{table}

\subsection{NCFF}
For every model and across all heads, the initial value is set as \(y_0=[2, 4]\) and \(\omega=1\).
\begin{table}[h]
  \caption{Scalability analysis for NCFF: Stiff regimes of the training heads.}\notag
  \centering
  \resizebox{0.7\linewidth}{!}{%
  \begin{tabular}{cccccccccccc}
  \toprule
    \textbf{Model} & \textbf{Head} & 1 & 2& 3& 4& 5& 6& 7& 8& 9& 10\\
    \midrule
    \(\textbf{1}\) &  \(\alpha\) & 1 & 2 & 3 & 4 & 5 & 6 & 7 & 8 & 9 & 10 \\
    \(\textbf{2}\) &     \(\alpha\) & 2 & 4 & 6 & 8 & 10 & 12 & 14 & 16 & 18 & 20 \\
    \(\textbf{3}\) &     \(\alpha\) & 3 & 6 & 9 & 12 & 15 & 18 & 21 & 24 & 27 & 30 \\
    \bottomrule 
  \end{tabular}}
\end{table}

\newpage
\subsection{Duffing}
For every model and across all heads, the initial value is set as \(y_0=[1, 0.5]\) and \(\beta=0.5\).
\begin{table}[h]
  \caption{Scalability analysis for Duffing: Stiff regimes of the training heads.}\notag
  \centering
  \resizebox{0.7\linewidth}{!}{%
  \begin{tabular}{cccccccccccc}
  \toprule
    \textbf{Model} & \textbf{Head} & 1 & 2& 3& 4& 5& 6& 7& 8& 9& 10\\
    \midrule
    \(\textbf{1}\) &  \(\alpha\) & 11 & 12 & 13 & 14 & 15 & 16 & 17 & 18 & 19 & 20 \\
    \(\textbf{2}\) &     \(\alpha\) & 12 & 14 & 16 & 18 & 20 & 22 & 24 & 26 & 28 & 30 \\
    \(\textbf{3}\) &     \(\alpha\) & 13 & 16 & 19 & 22 & 25 & 28 & 31 & 34 & 37 & 40 \\
    \bottomrule 
  \end{tabular}}
\end{table}
\subsection{AR}
For every model and across all heads, the initial value is set as \(y_0^{max}=1\).
\begin{table}[!h]
  \caption{Scalability analysis for AR: Stiff regimes of the training heads.}\notag
  \centering
  \resizebox{0.35\linewidth}{!}{%
  \begin{tabular}{cccccccccccc}
  \toprule
    \textbf{Model} & \textbf{Head} & 1 & 2& 3& 4\\
    \midrule
    \(\textbf{1}\) &  \(\alpha\) & 1 & 2 & 3 & 4 \\
    \(\textbf{2}\) &     \(\alpha\) & 3 & 6 & 9 & 12 \\
    \bottomrule 
  \end{tabular}}
\end{table}

\section{Reparametrization Analysis: Training Heads }\label{app:I}
We present the training head configurations of MH-PINN in a low-stiff regime of stiff-parameterized ODEs and PDE with varying initial conditions or forcing function. For OHO: \(y_{1_0}\) and \(y_{2_0}\!\! \in \! [0, 5]\), Duffing: \(y_{1_0} \!\! \in \! [0.5, 2]\) and \(y_{2_0} \!\! \in \! [0, 1]\), AR equation:  \(y_0^{max} \!\! \in \! [0.5, 2]\), NCFF: \(\omega \! \in \! [0, \pi]\) uniformly sample. Values are given with 2 C.S.

\begin{table}[h]
  \caption{Reparametrization Analysis: MH-PINN training heads}\notag
  \centering
  \resizebox{0.95\linewidth}{!}{%
  \begin{tabular}{lccccccccccc}
  \toprule
    & \textbf{Head} & 1 & 2& 3& 4& 5& 6& 7& 8& 9& 10\\
    \midrule
    \multirow{3}{*}{\textbf{OHO}} &     \(\alpha\) & 2 & 4 & 6 & 8 & 10 & 12 & 14 & 16 & 18 & 20 \\
    & \(y_{1_0}\) & 0.59 & 1.1 & 1.4 & 2.9 & 3.7 & 4.2 & 4.1 & 0.30 & 2.2 & 1.7 \\
    & \(y_{2_0}\) & 1.5 & 3.3 & 3.7 & 3.3 & 3.1 & 4.7 & 0.59 & 3.9 & 0.86 & 1.3 \\
    \midrule
    \multirow{3}{*}{\textbf{NCFF}} &     \(\alpha\) & 2 & 4 & 6 & 8 & 10 & 12 & 14 & 16 & 18 & 20 \\
    & \(y_0\) & [2, 4] & [2, 4] & [2, 4] & [2, 4] & [2, 4] & [2, 4] & [2, 4] & [2, 4] & [2, 4] & [2, 4] \\
    &     \(\omega\) & 2.3 & 0.49 & 0.059 & 0.12 & 0.98 & 3.1 & 0.82 & 1.1 & 0.67 & 2.7 \\
    \midrule
    \multirow{4}{*}{\textbf{Duffing}} &     \(\alpha\) & 13 & 16 & 19 & 22 & 25 & 28 & 31 & 34 & 37 & 40 \\
    & \(y_{1_0}\) & 1.9 & 1.4 & 1.7 & 0.62 & 0.58 & 1.1 & 0.63 & 0.50 & 1.7 & 1.4\\
    & \(y_{2_0}\) & 0.59 & 0.62 & 0.19 & 0.32 & 0.34 & 0.81 & 0.51 & 0.26 & 0.81 & 0.43\\
    &     \(\beta\) & 0.5 & 0.5 & 0.5 & 0.5 & 0.5 & 0.5 & 0.5 & 0.5 & 0.5 & 0.5 \\
    \midrule
    \multirow{2}{*}{\textbf{AR}} &     \(\alpha\) & 3 & 6 & 8 & 12 & - & - & - & - & - & - \\
    & \(y_0^{max}\!\!\!\!\) & 0.50 & 1.0 & 1.5 & 2.0 & - & - & - & - & - & -\\
    \bottomrule 
  \end{tabular}}
\end{table}

\newpage
\section{Performance Analysis: Training and Transfer Learning Results}\label{app:J}
In this section, we give more figures about training and transfer learning.
For each equation, we provide the results of a run, including the training loss curve, training head output, transfer learning solutions for \(\alpha \in [10, 200]\) and the associated absolute error. 

\subsection{OHO}
Here are the performance analysis, training and transfer results of the OHO equation.
\begin{figure}[hbt!]
  \centering
\includegraphics[width=0.95\linewidth]{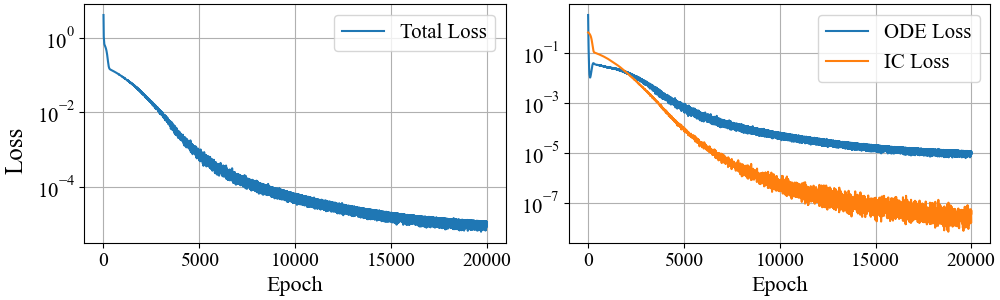}
  \caption{Training loss, along with the loss values for ODE and initial condition (IC), across epochs.} 
  \label{fig:6}
\end{figure}
\FloatBarrier

\begin{figure}[hbt!]
  \centering
\includegraphics[width=1\linewidth]{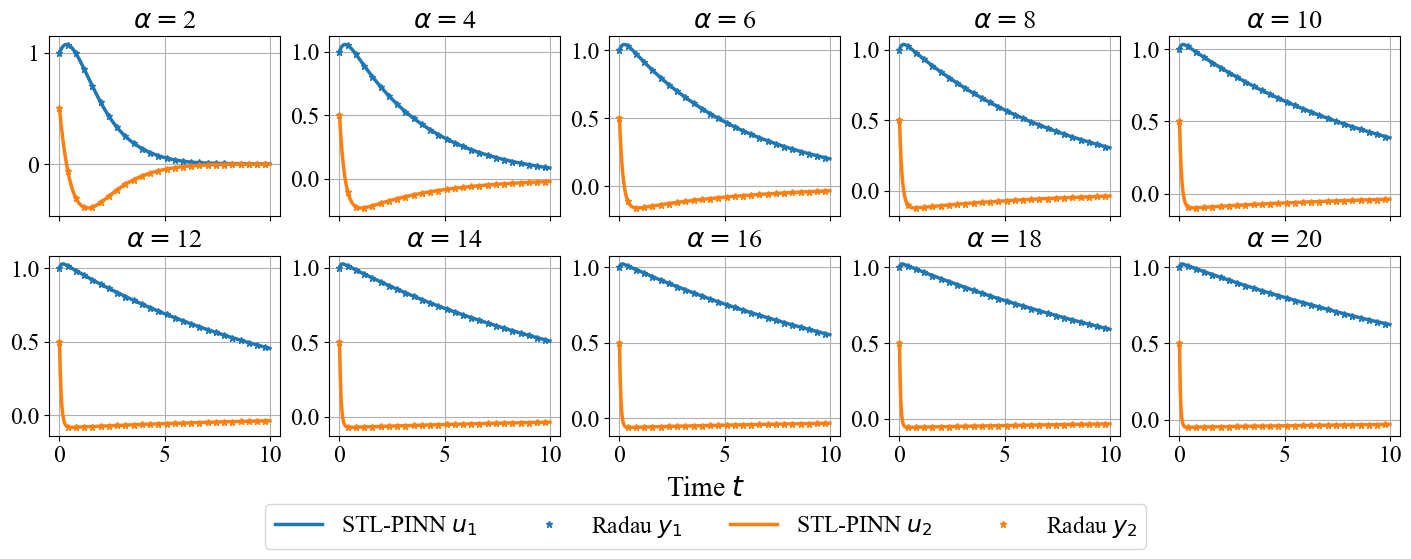}
  \caption{MH-PINN outcomes after training, along with the Radau numerical solution across all training heads.} 
  \label{fig:7}
\end{figure}
\FloatBarrier

From Figure \ref{fig:6} and Figure \ref{fig:7}, we see that the network effectively learned the low-stiff solutions, as shown by the decreasing loss. Additionally, the numerical solutions closely align with MH-PINN solutions across all heads.

\begin{figure}[hbt!]
  \centering
\includegraphics[width=1\linewidth]{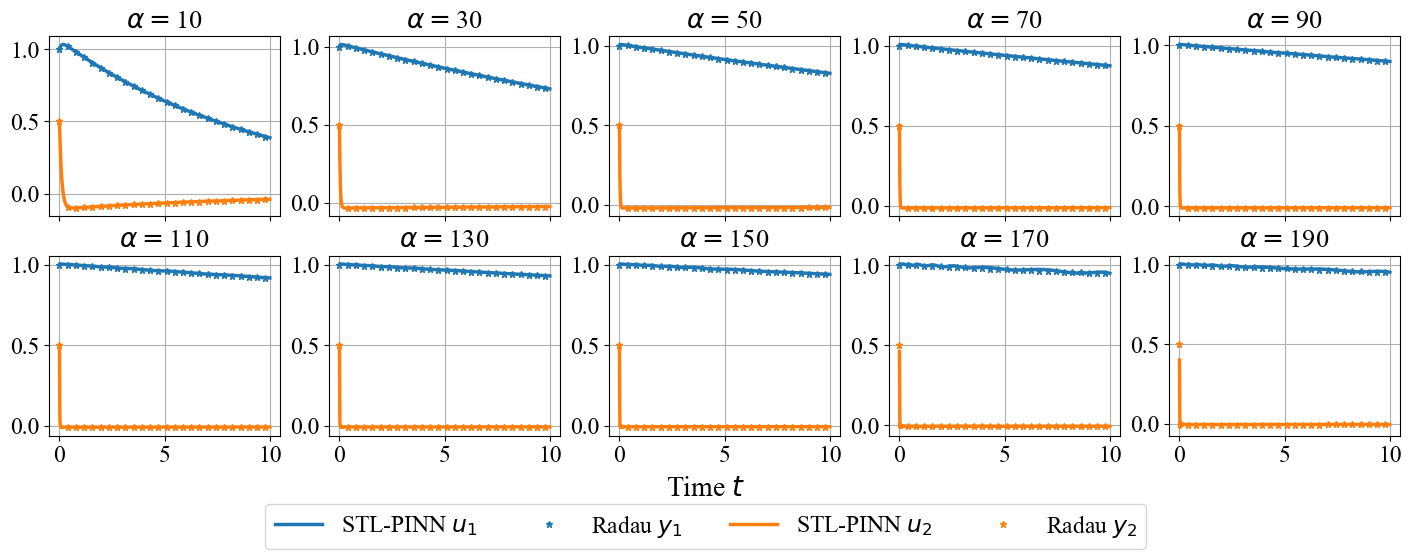}
  \caption{STL-PINN outcomes after transfer learning for \(\alpha\!\in\![10, 200]\), along with the Radau numerical solution.} 
  \label{fig:8}
\end{figure}
\FloatBarrier

\begin{figure}[hbt!]
  \centering
\includegraphics[width=1\linewidth]{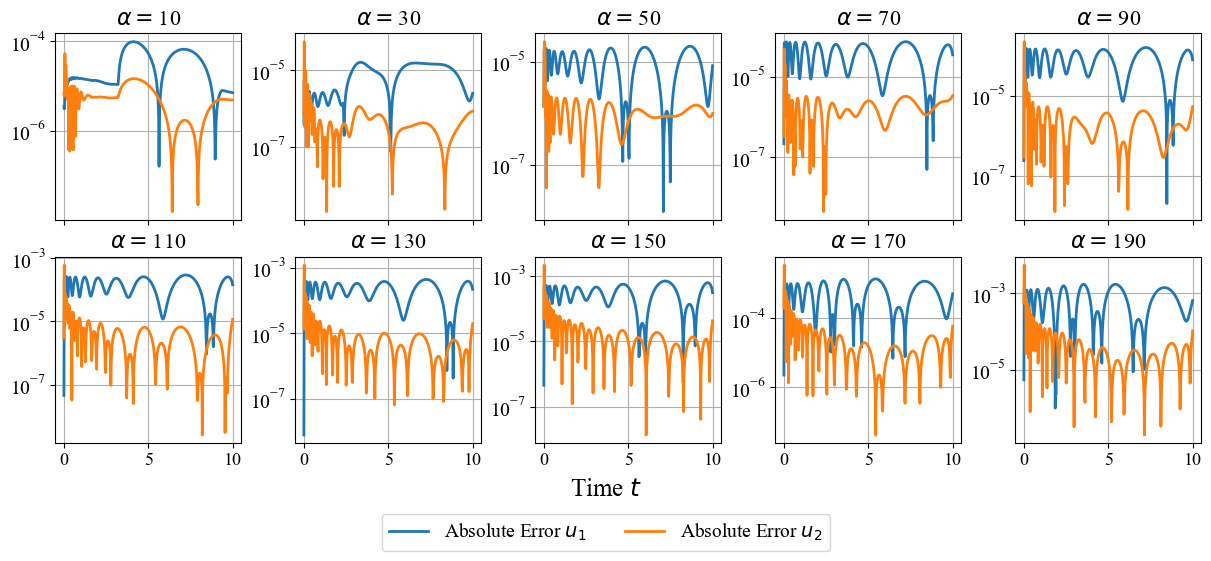}
  \caption{Absolute error of STL-PINN with respect to the Radau solution for \(\alpha \!\in\! [10, 200]\).} 
  \label{fig:9}
\end{figure}
\FloatBarrier

Figure \ref{fig:8} and Figure \ref{fig:9} illustrate that STL-PINN can compute accurate solutions even in high-stiff regimes. It captures stiffness behaviors well, such as the transient phase near the initial condition of \(u_2\). However, errors increase with stiffness as we move further away from the low-stiff training regime.

\newpage
\subsection{NCFF}
Here are the performance analysis, training and transfer results of the NCFF equation.
\begin{figure}[hbt!]
  \centering
\includegraphics[width=0.95\linewidth]{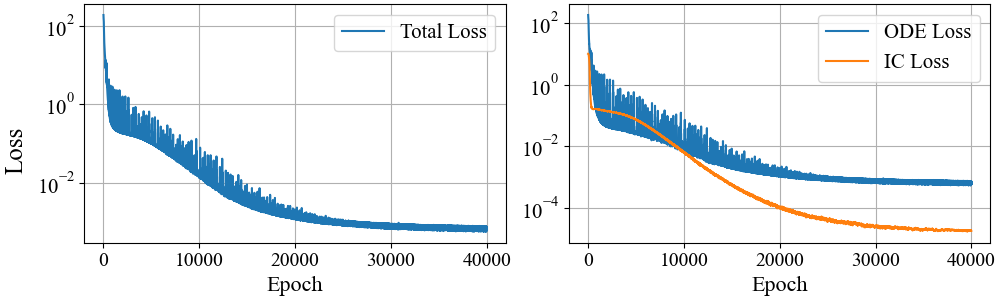}
  \caption{Training loss, along with the loss values for ODE and initial condition (IC), across epochs.} 
  \label{fig:10}
\end{figure}
\FloatBarrier
\begin{figure}[hbt!]
  \centering
\includegraphics[width=1\linewidth]{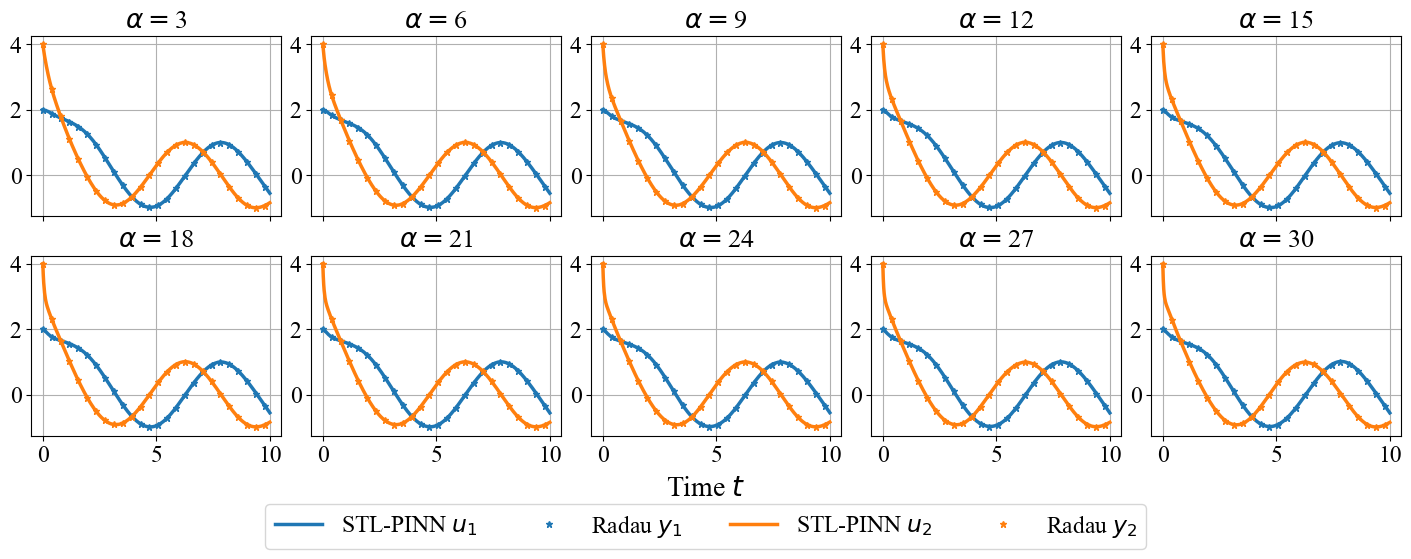}
  \caption{MH-PINN outcomes after training, along with the Radau numerical solution across all training heads.} 
  \label{fig:11}
\end{figure}
\FloatBarrier

From Figure \ref{fig:10} and Figure \ref{fig:11}, we see that the network effectively learned the low-stiff solutions, as shown by the decreasing loss. Additionally, the numerical solutions closely align with MH-PINN solutions across all heads.

\begin{figure}[hbt!]
  \centering
\includegraphics[width=1\linewidth]{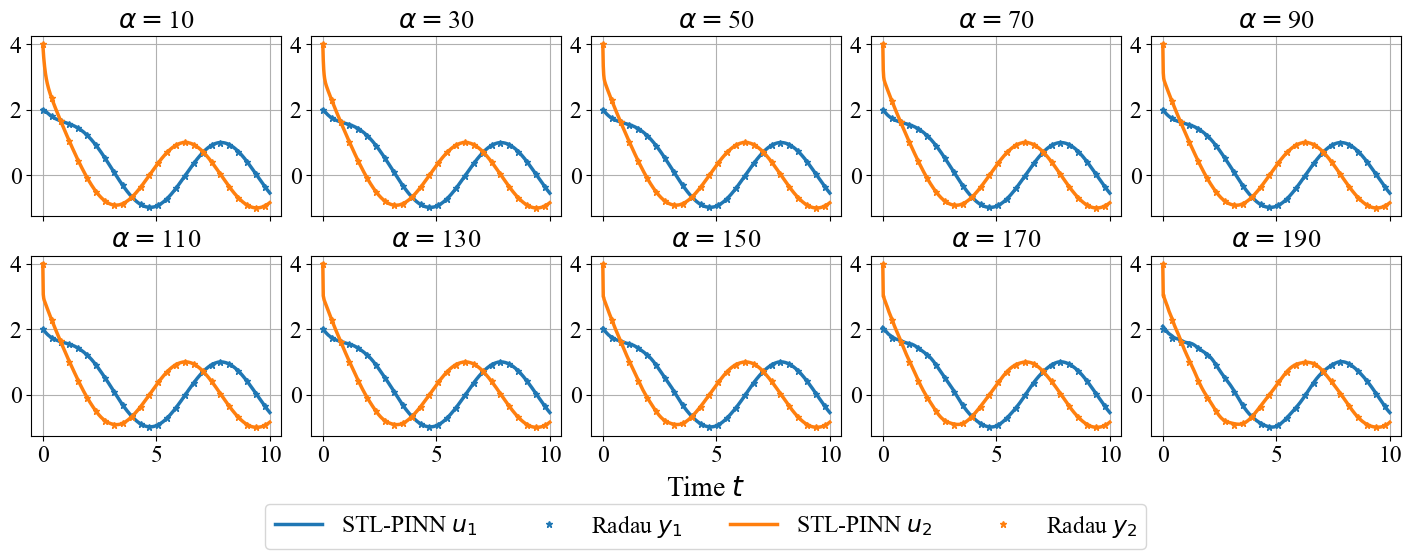}
  \caption{STL-PINN outcomes after transfer learning for \(\alpha\!\in\![10, 200]\), along with the Radau numerical solution.} 
  \label{fig:12}
\end{figure}
\FloatBarrier
\begin{figure}[hbt!]
  \centering
\includegraphics[width=1\linewidth]{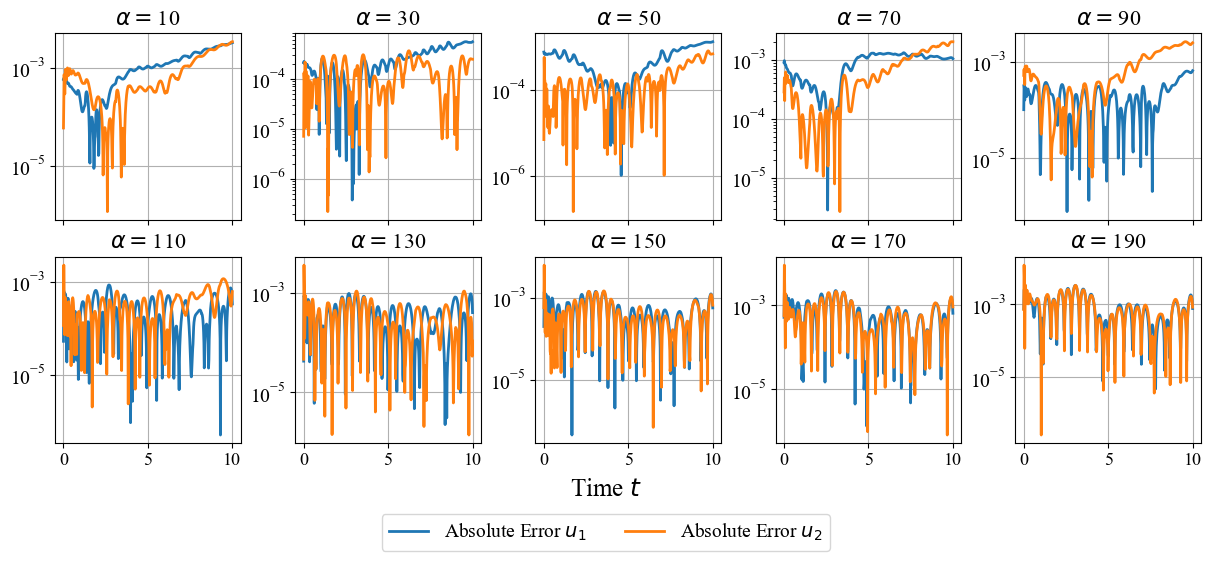}
  \caption{Absolute error of STL-PINN with respect to the Radau solution for \(\alpha \in [10, 200]\).} 
  \label{fig:13}
\end{figure}
\FloatBarrier

Figure \ref{fig:12} and Figure \ref{fig:13} illustrate that STL-PINN can compute accurate solutions even in high-stiff regimes. It captures stiffness behaviors well, such as the transient phase near the initial condition of \(u_2\). However, errors increase with stiffness as we move further away from the low-stiff training regime.

\newpage
\subsection{Duffing}
Here are the performance analysis, training and transfer results of the Duffing equation.
\begin{figure}[hbt!]
  \centering
\includegraphics[width=0.95\linewidth]{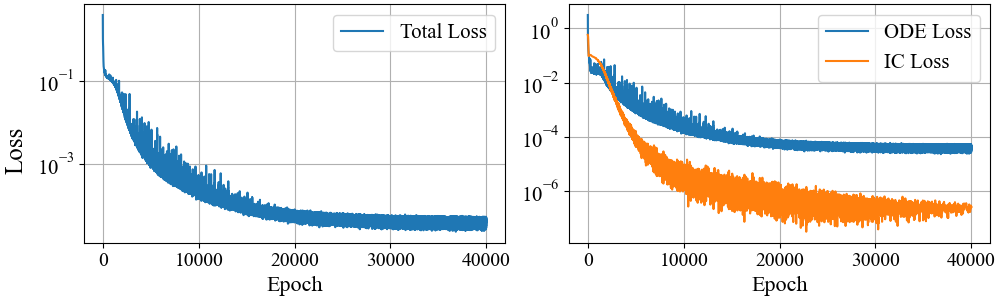}
  \caption{Training loss, along with the loss values for ODE and initial condition (IC), across epochs.} 
  \label{fig:14}
\end{figure}
\FloatBarrier
\begin{figure}[hbt!]
  \centering
\includegraphics[width=1\linewidth]{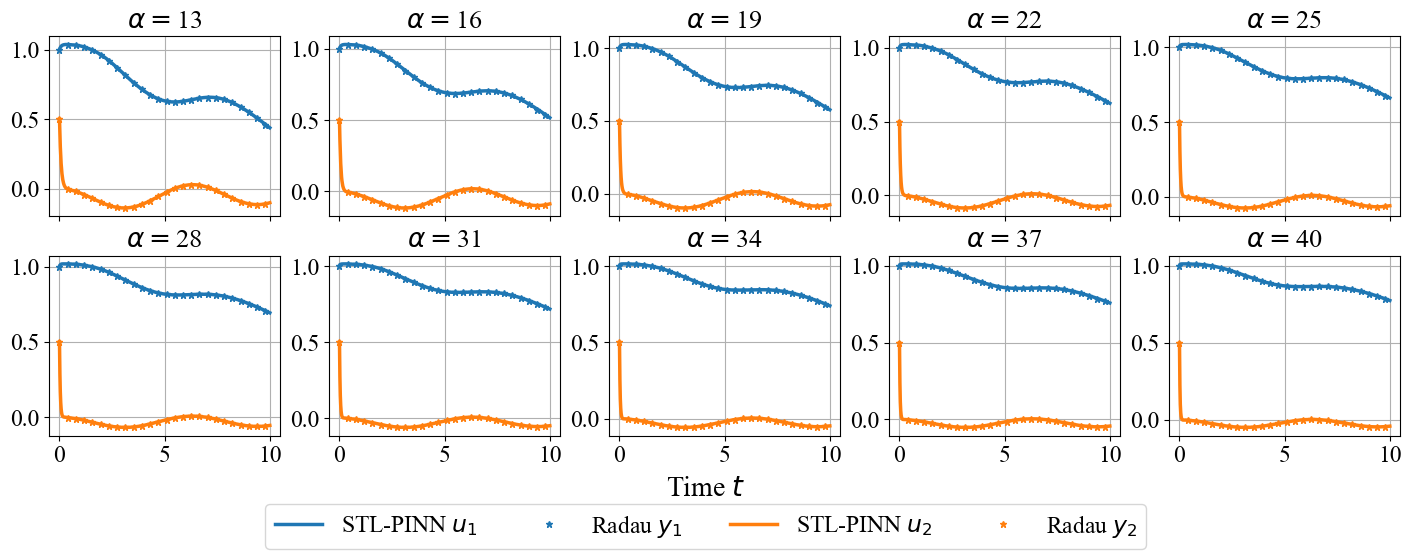}
  \caption{MH-PINN outcomes after training, along with the Radau numerical solution across all training heads..} 
  \label{fig:15}
\end{figure}
\FloatBarrier

For the Duffing equation, MH-PINNs is trained on its linear form, as the nonlinear part is then approximated during transfer learning with the perturbation expansion. Here, the training results are given for the linear form of the equation. From Figure \ref{fig:14} and Figure \ref{fig:15}, it's clear that the network effectively learned the low-stiff solutions, as shown by the decreasing loss. Additionally, the numerical solutions closely align with MH-PINN solutions across all heads.

\begin{figure}[hbt!]
  \centering
\includegraphics[width=1\linewidth]{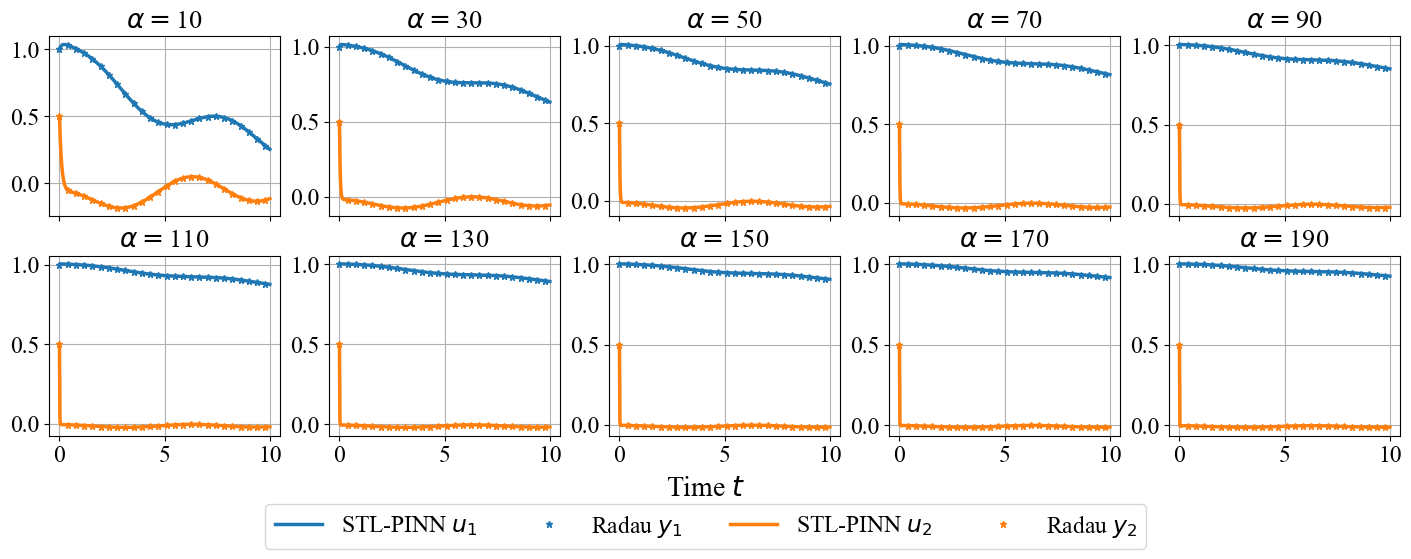}
  \caption{STL-PINN outcomes after transfer learning for \(\alpha\!\in\![10, 200]\), along with the Radau numerical solution.} 
  \label{fig:16}
\end{figure}
\FloatBarrier
\begin{figure}[hbt!]
  \centering
\includegraphics[width=1\linewidth]{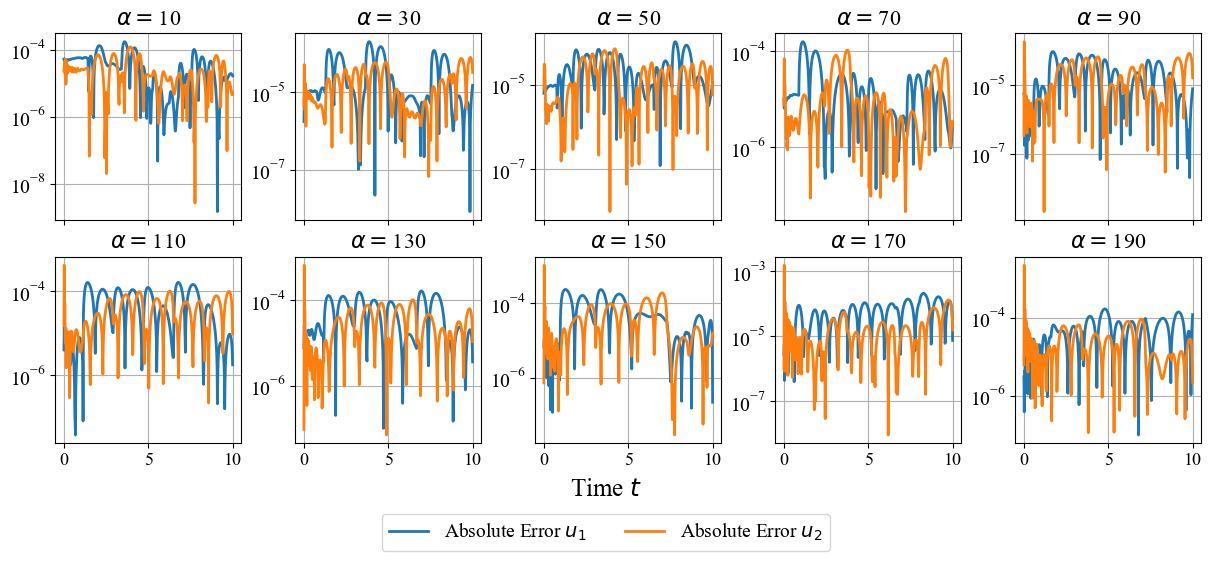}
  \caption{Absolute error of STL-PINN with respect to the Radau solution for \(\alpha\!\in\![10, 200]\).} 
  \label{fig:17}
\end{figure}
\FloatBarrier

Here, we present the transfer learning results on the nonlinear form using the perturbation expansion. Figure \ref{fig:16} and Figure \ref{fig:17} illustrate that STL-PINN can compute accurate solutions even in high-stiff regimes. It also effectively approximates the nonlinearity through the perturbation expansion. It captures stiffness behaviors well, such as the transient phase near the initial condition of \(u_2\). However, errors increase with stiffness as we move further away from the low-stiff training regime.

\newpage
\subsection{AR}
Here are the performance analysis, training and transfer results of the AR equation.
\begin{figure}[hbt!]
  \centering
\includegraphics[width=0.95\linewidth]{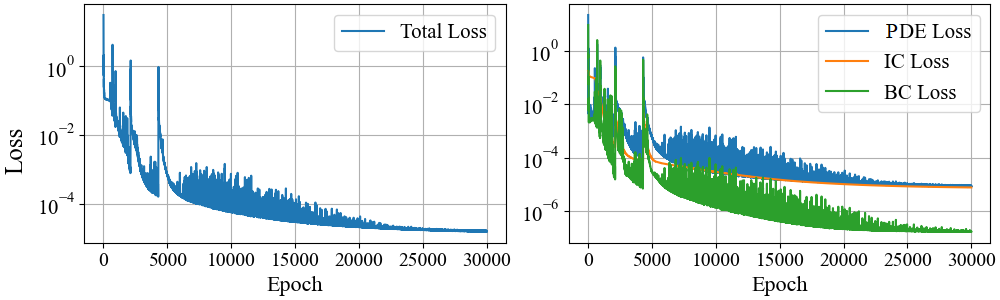}
  \caption{Training loss, along with the loss values for ODE and initial condition (IC) and boundary condition (BC) across epochs.} 
  \label{fig:18}
\end{figure}
\FloatBarrier
\begin{figure}[hbt!]
  \centering
\includegraphics[width=1\linewidth]{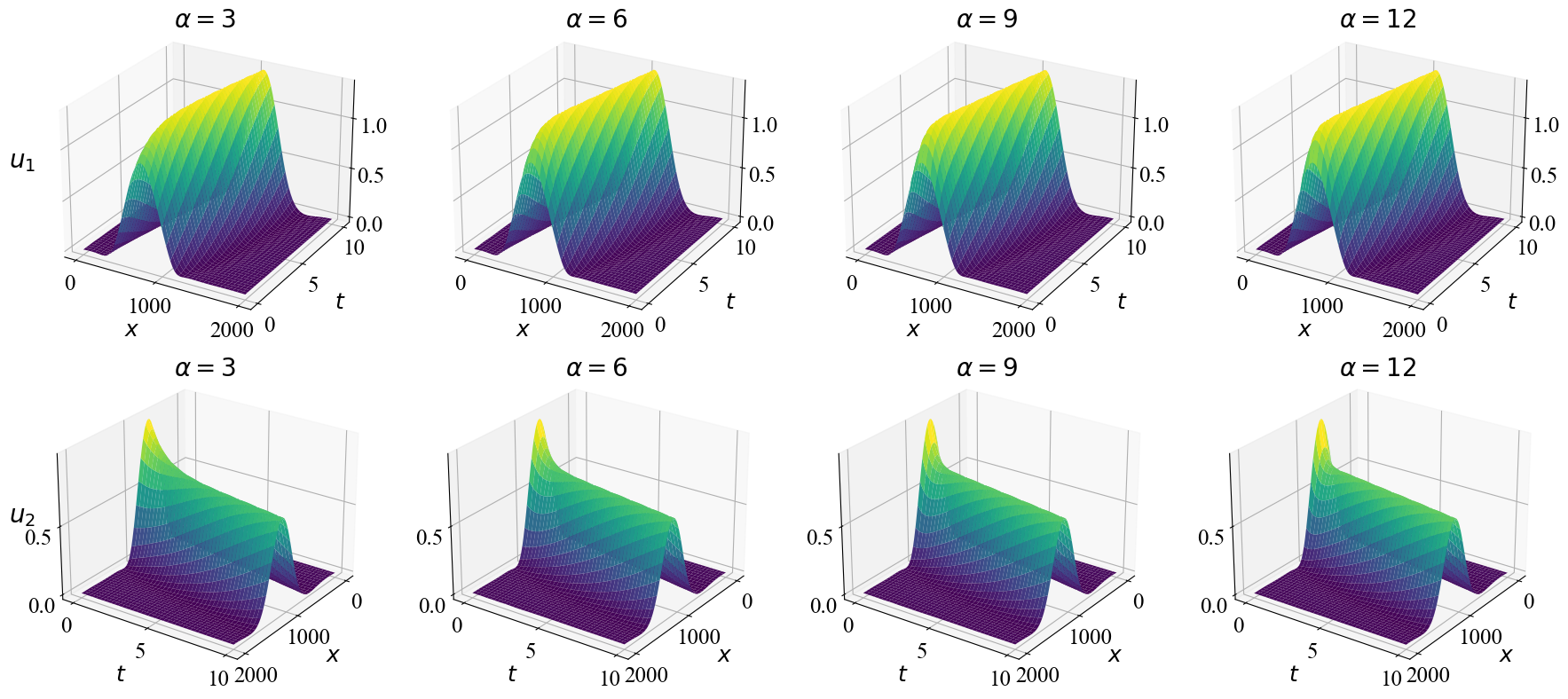}
  \caption{MH-PINN outcomes after training across all training heads.} 
  \label{fig:19}
\end{figure}

From Figure \ref{fig:18} and Figure \ref{fig:19}, we see that the network effectively learned the low-stiff solutions, as shown by the decreasing loss even for the boundary and initial conditions. We observe that \(u_1\) and \(u_2\) correctly reach equilibrium after a relatively slow transient phase due to the low stiffness regimes.

\begin{figure}[hbt!]
  \centering
\includegraphics[width=1\linewidth]{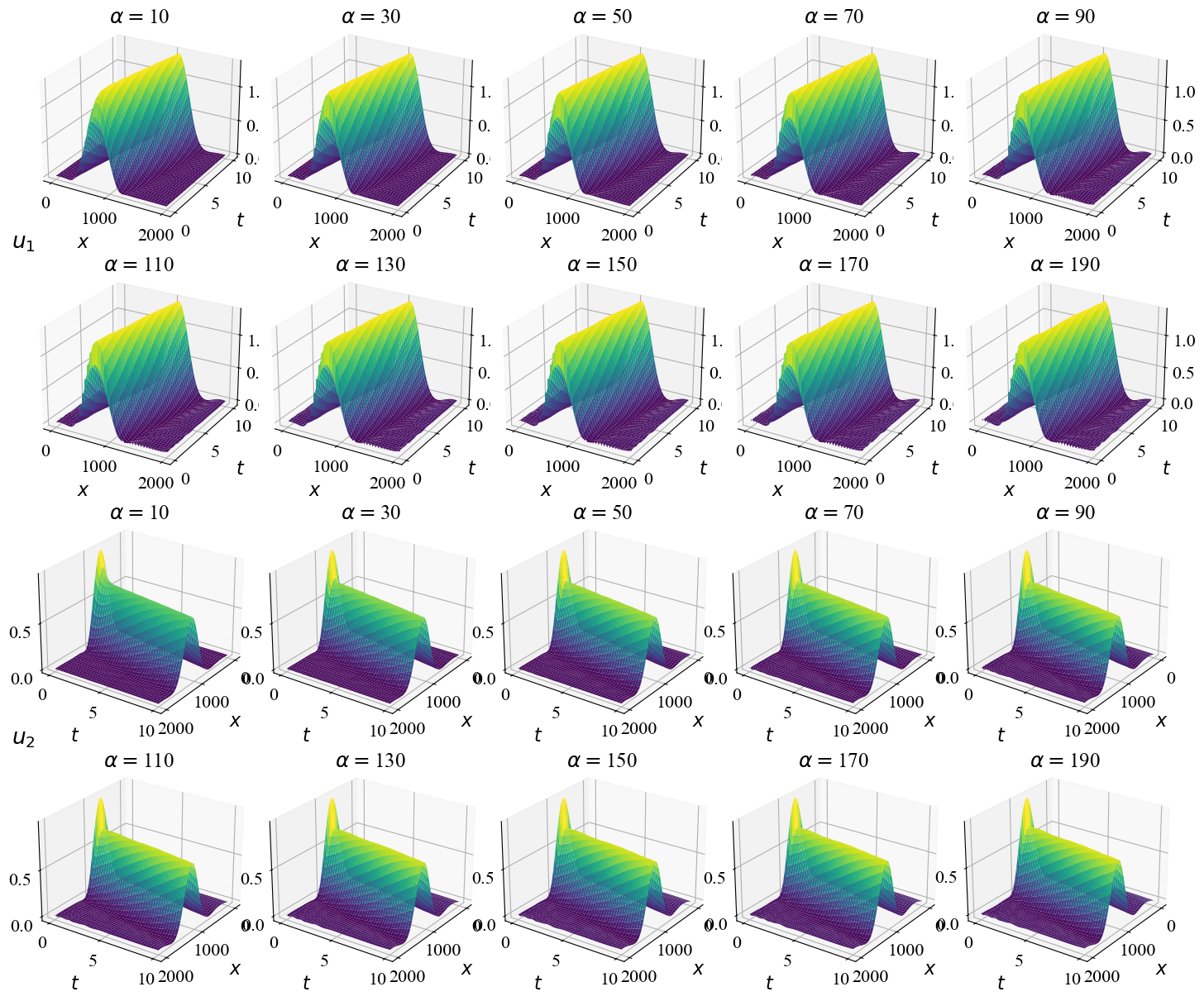}
  \caption{STL-PINN outcomes after transfer learning for \(\alpha\!\in\![10, 200]\).} 
  \label{fig:20}
\end{figure}
\FloatBarrier
\begin{figure}[hbt!]
  \centering
\includegraphics[width=1\linewidth]{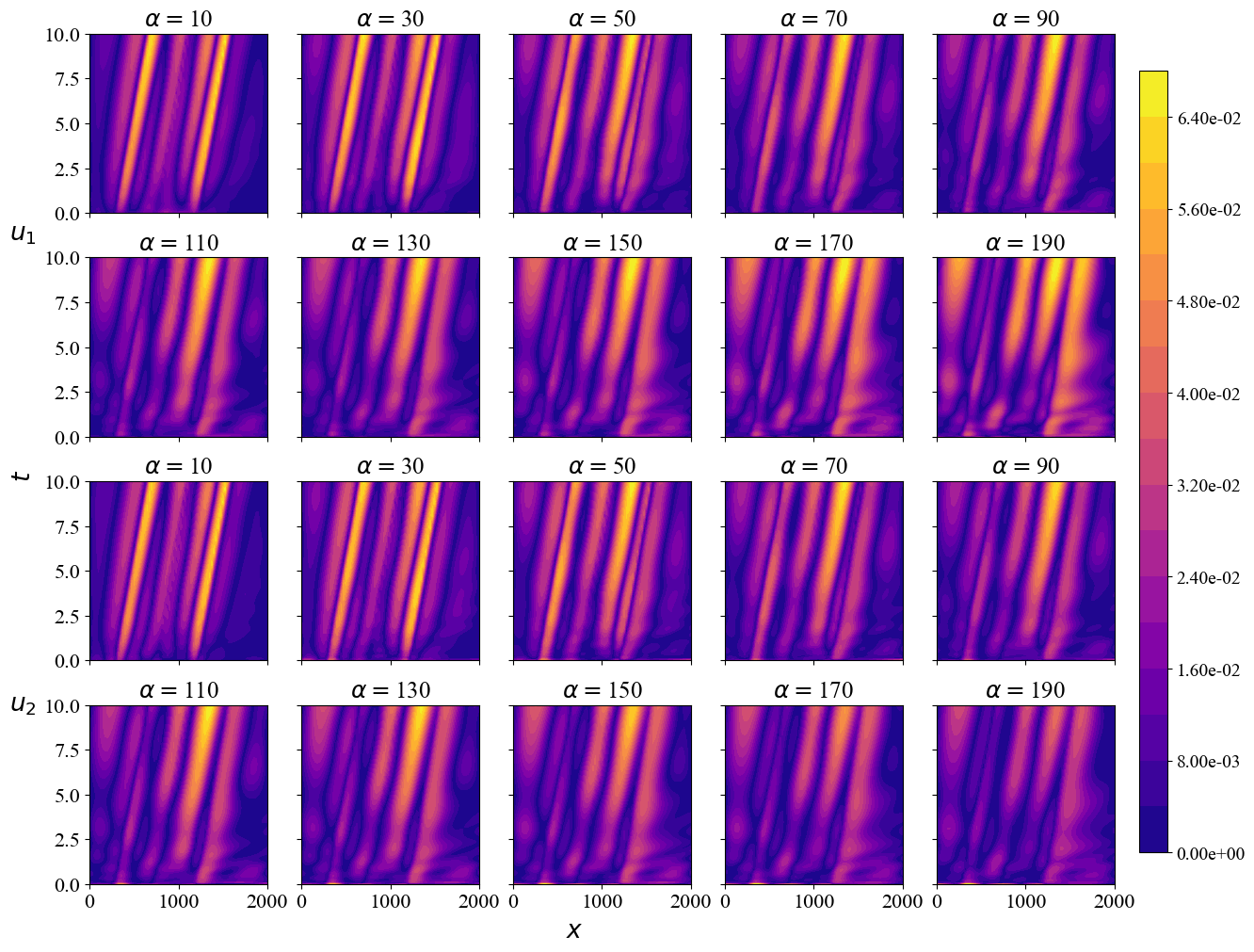}
  \caption{Absolute error of STL-PINN with respect to the LW-Radau solution for \(\alpha \in [10, 200]\).} 
  \label{fig:21}
\end{figure}
\FloatBarrier
Figure \ref{fig:20} and Figure \ref{fig:21} illustrate that STL-PINN can compute accurate PDEs solutions even in high-stiff regimes. Transient phases are now very rapid, with solutions reaching equilibrium quickly. However, errors increase with stiffness as we move further away from the low-stiff training regime.

\section{Performance Analysis: More Metrics
}\label{app:more_mertics}

\begin{figure}[hbt!]
  \centering
\includegraphics[width=0.95\linewidth]{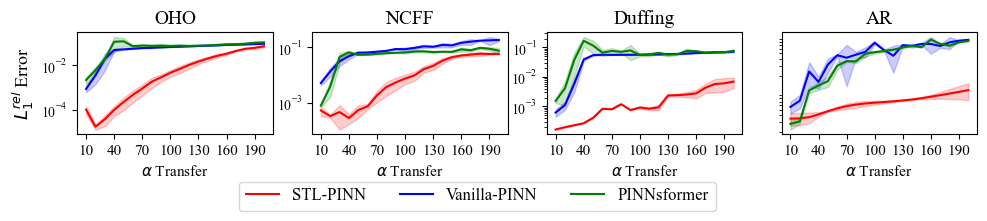}
  \caption{\(L_1\) relative error over vanilla PINN, PINNsFormer and STL-PINN solving stiff-parameterized ODEs-PDEs as stiffness regime increases, with \(\alpha \in [10, 200]\). The reported errors are the average of three separate runs, along with 90\% confidence intervals.} 
  \label{fig:L1rel}
\end{figure}

\begin{figure}[hbt!]
  \centering
\includegraphics[width=0.95\linewidth]{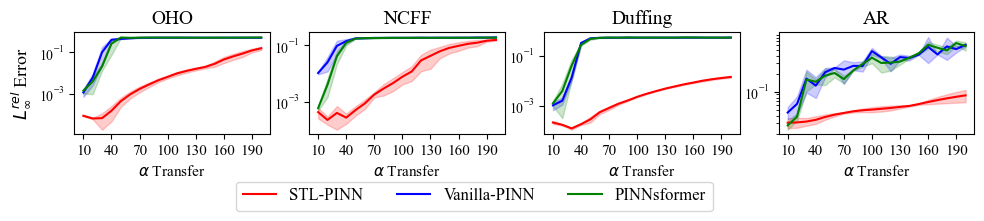}
  \caption{\(L_{\infty}\) relative error of vanilla PINN, PINNsFormer and STL-PINN solving stiff-parameterized ODEs-PDEs as stiffness regime increases, with \(\alpha \in [10, 200]\). The reported errors are the average of three separate runs, along with 90\% confidence intervals.} 
  \label{fig:Linfrel}
\end{figure}

Here, Figure \ref{fig:L1rel} and \ref{fig:Linfrel} show \(L_1\) relative error and \(L_{\infty}\) relative error of the performance analysis. STL-PINN consistently outperforms vanilla PINNs and PINNsFormer methods across all $\alpha$ values.

\newpage
\section{Scalability Analysis: More Metrics
}\label{app:M}
\begin{figure}[hbt!]
  \centering
\includegraphics[width=0.95\linewidth]{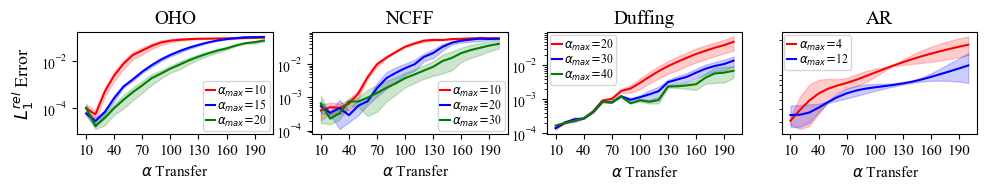}
  \caption{\(L_1\) relative error of STL-PINN over increasing ranges of \(\alpha\) during training, where \(\alpha_{max}\) denotes the maximum value. The x-axis is the \(\alpha \) transfer range. The reported errors are the average of three separate runs, along with 90\% confidence intervals.} 
  \label{fig:22}
\end{figure}
\FloatBarrier

\begin{figure}[hbt!]
  \centering
\includegraphics[width=0.95\linewidth]{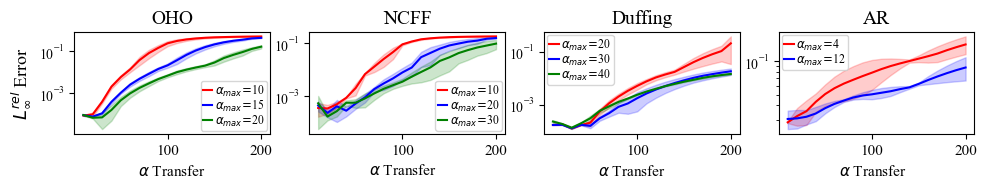}
  \caption{\(L_{\infty}\) relative error of STL-PINN over increasing ranges of \(\alpha\) during training, where \(\alpha_{max}\) denotes the maximum value. The x-axis is the \(\alpha \) transfer range. The reported errors are the average of three separate runs, along with 90\% confidence intervals.} 
  \label{fig:23}
\end{figure}
\FloatBarrier

Here, Figure \ref{fig:22} and \ref{fig:23} show \(L_1\) relative error and \(L_{\infty}\) relative error of the scalability analysis. As mentioned in the result section, it demonstrates the scalability of our method: the more we train in stiff regimes, the more we extend the range of transfer learning to even stiffer regimes.

\newpage
\section{Reparametrization Analysis: Training Results
}\label{app:K}
In this section, we give figures about training results of the reparametrization analysis.
For each equation, we provide the training outcomes including training loss curve and training head output.

\subsection{OHO}
\begin{figure}[hbt!]
  \centering
\includegraphics[width=0.95\linewidth]{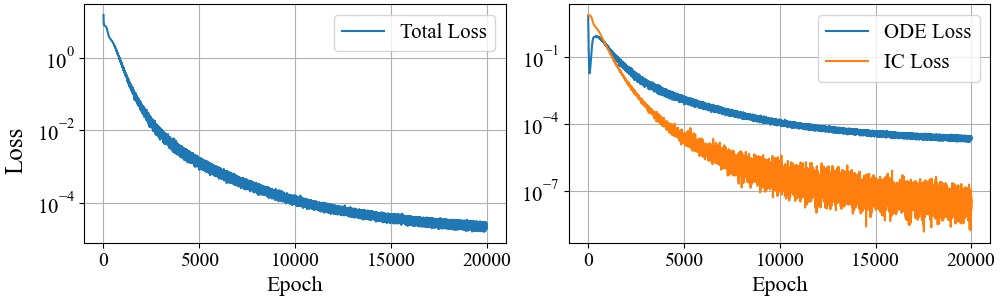}
  \caption{Training loss of the reparametrization analysis, along with the loss values for ODE and initial condition (IC), across epochs.} 
  \label{fig:24}
\end{figure}
\FloatBarrier
\begin{figure}[hbt!]
  \centering
\includegraphics[width=1\linewidth]{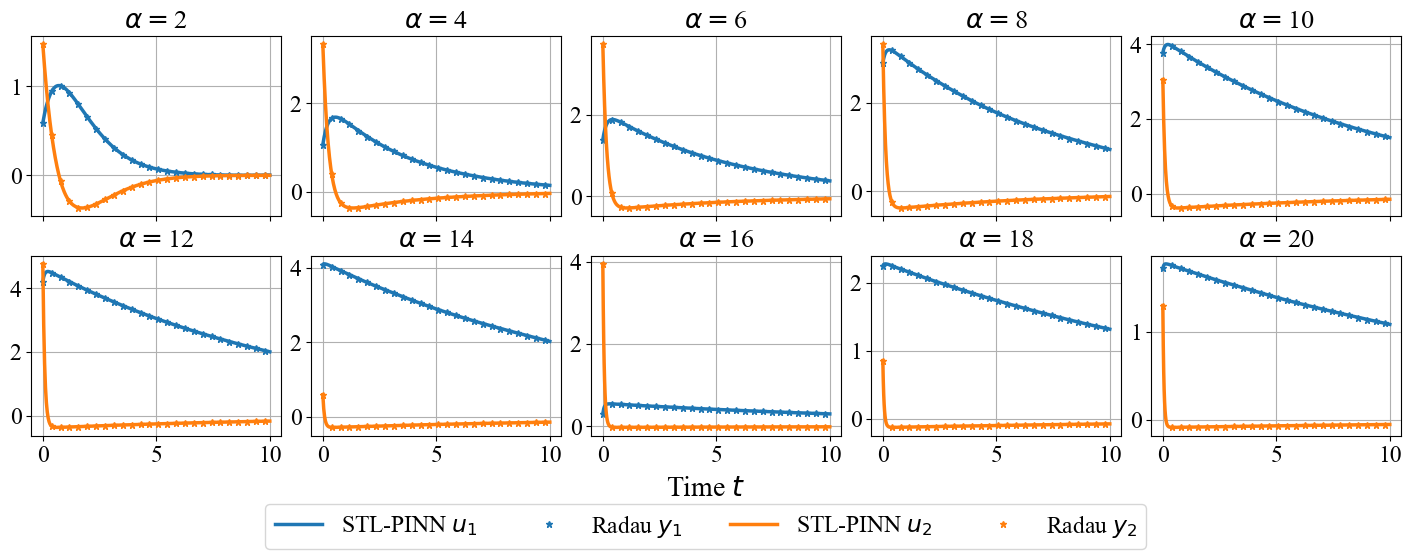}
  \caption{MH-PINN outcomes after training with different initial conditions, along with the Radau numerical solution over the 10 heads.} 
  \label{fig:25}
\end{figure}
\FloatBarrier

Figures \ref{fig:24} and \ref{fig:25} show that the network effectively learned solutions with various initial conditions, as shown by the decreasing loss. Additionally, the numerical solutions closely align with MH-PINN solutions across all heads, successfully encoding the different initial conditions.

\newpage
\subsection{NCFF}
\begin{figure}[hbt!]
  \centering
\includegraphics[width=0.95\linewidth]{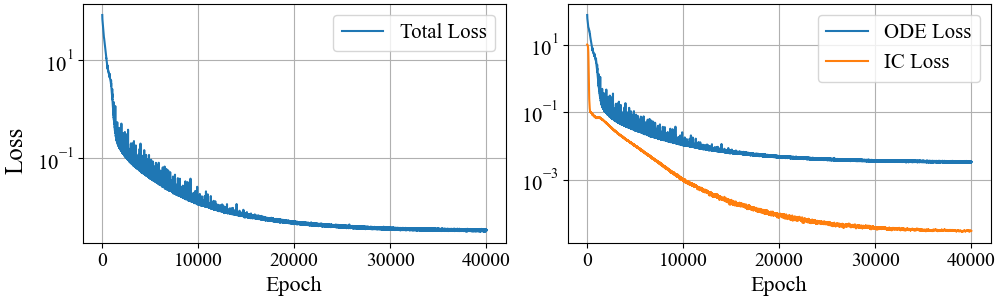}
  \caption{Training loss of the reparametrization analysis, along with the loss values for ODE and initial condition (IC), across epochs.} 
  \label{fig:26}
\end{figure}
\FloatBarrier
\begin{figure}[hbt!]
  \centering
\includegraphics[width=1\linewidth]{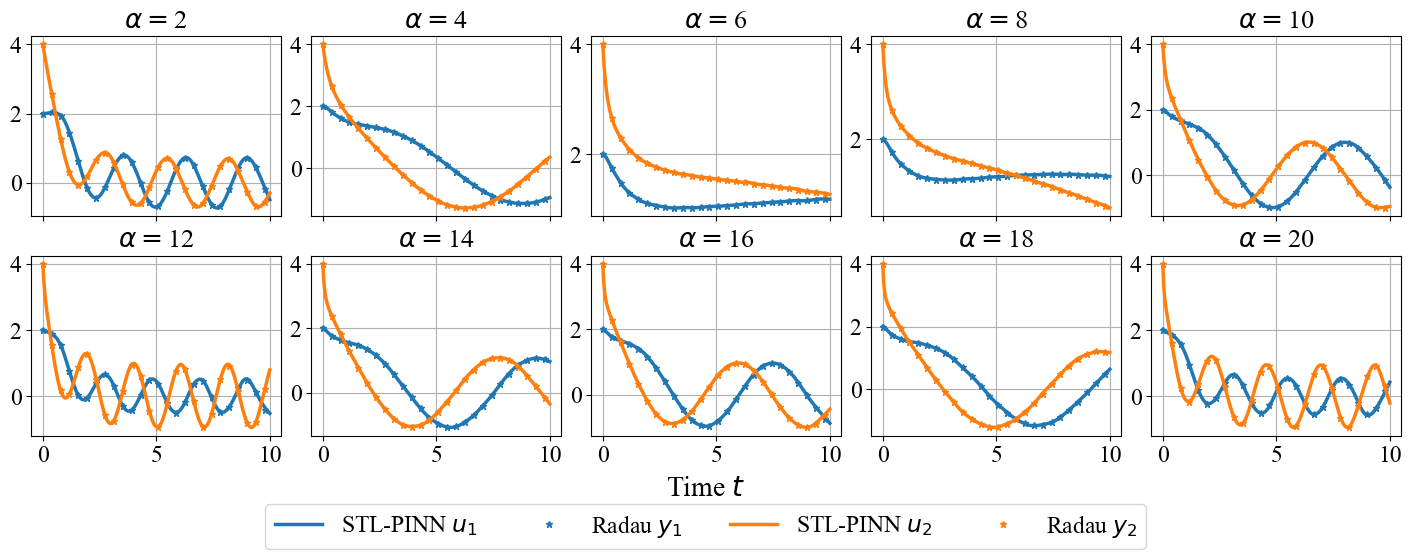}
  \caption{MH-PINN outcomes after training with different forcing functions, along with the Radau numerical solution over the 10 heads.} 
  \label{fig:27}
\end{figure}
\FloatBarrier

Figures \ref{fig:26} and \ref{fig:27} show that the network effectively learned solutions with various forcing functions, as shown by the decreasing loss. Additionally, the numerical solutions closely align with MH-PINN solutions across all heads, successfully encoding the different forcing  functions.

\newpage
\subsection{Duffing}
\begin{figure}[hbt!]
  \centering
\includegraphics[width=0.95\linewidth]{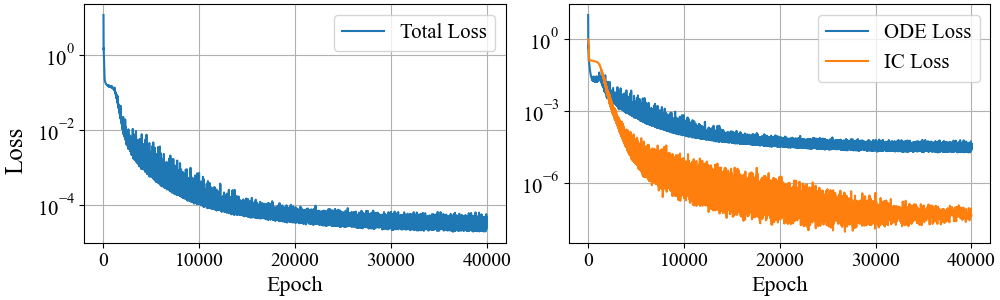}
  \caption{Training loss of the reparametrization analysis, along with the loss values for ODE and initial condition (IC), across epochs.} 
  \label{fig:28d}
\end{figure}
\FloatBarrier
\begin{figure}[hbt!]
  \centering
\includegraphics[width=1\linewidth]{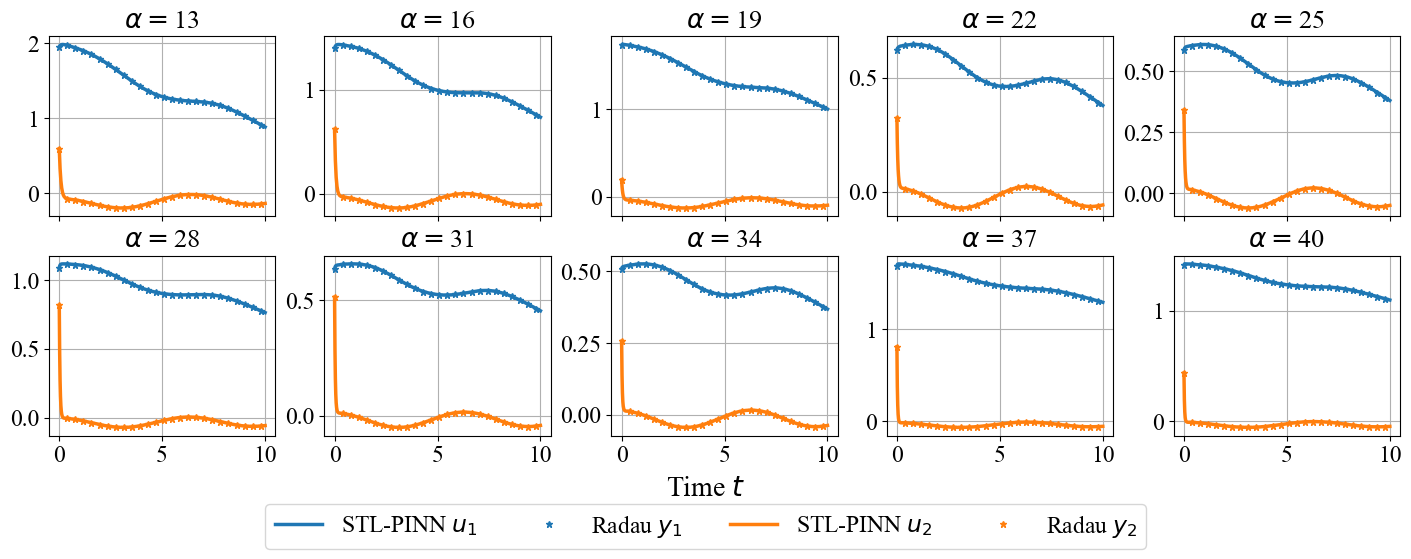}
  \caption{MH-PINN outcomes after training with different initial condition, along with the Radau numerical solution over the 4 heads.} 
  \label{fig:29d}
\end{figure}
\FloatBarrier

For the Duffing equation, MH-PINNs are trained on its linear form, as the nonlinear part is then approximated during transfer learning with the perturbation expansion. Here, the training results are given for the linear form of the equation. Figures \ref{fig:28d} and \ref{fig:29d} show that the network effectively learned solutions with various initial conditions, as shown by the decreasing loss. Additionally, the numerical solutions closely align with MH-PINN solutions across all heads, successfully encoding the different initial conditions.

\newpage
\subsection{AR}
\begin{figure}[hbt!]
  \centering
\includegraphics[width=0.95\linewidth]{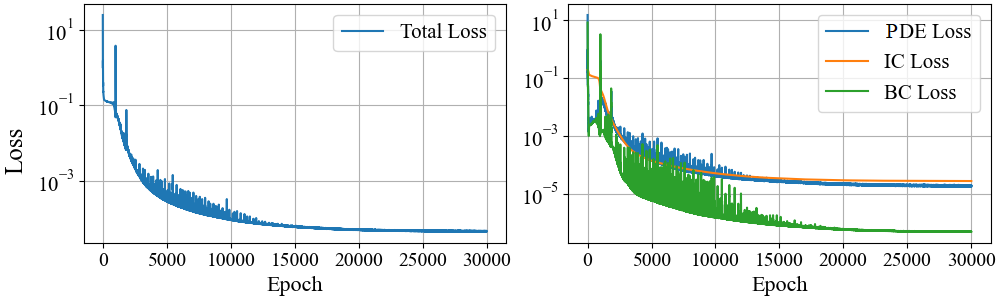}
  \caption{Training loss of the reparametrization analysis, along with the loss values for ODE, initial condition (IC) and boundary condition (BC), across epochs.} 
  \label{fig:28}
\end{figure}
\FloatBarrier
\begin{figure}[hbt!]
  \centering
\includegraphics[width=1\linewidth]{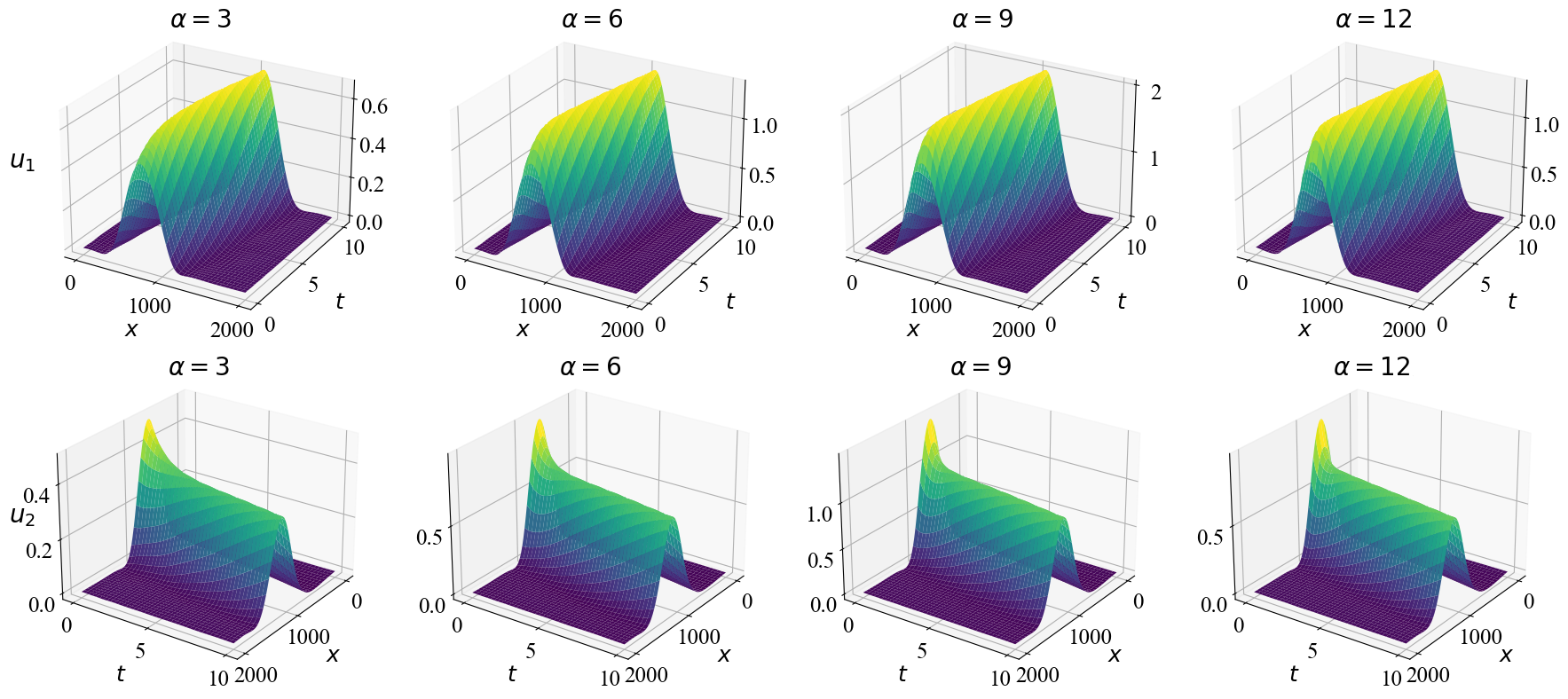}
  \caption{MH-PINN outcomes after training with different initial condition, over the 4 heads.} 
  \label{fig:29}
\end{figure}
\FloatBarrier

From Figure \ref{fig:28} and Figure \ref{fig:29}, we see that the network effectively learned the low-stiff solutions with various initial conditions, as shown by the decreasing loss.

\end{document}